\pdfoutput=1
\documentclass{article}
\usepackage{microtype}
\usepackage{graphicx}
\usepackage{subcaption}
\usepackage{booktabs}
\usepackage{hyperref}

  \usepackage[preprint]{icml2026}
  \hypersetup{pdfsubject={}}

\usepackage{amsmath}
\usepackage{amssymb}
\usepackage{mathtools}
\usepackage{amsthm}
\usepackage[capitalize,noabbrev]{cleveref}
\usepackage{siunitx}
\usepackage{xspace}

\renewcommand{\Re}{\mathrm{Re}} %
\newcommand{\Rel}{\mathrm{Re}_\lambda}
\newcommand{\grid}[1]{#1^3}

\newcommand{\resIsoMainU}{6.54}
\newcommand{\resIsoMainP}{9.88}

\newcommand{\resIsoMainParams}{9.88}

\newcommand{\resIsoMainBelowTransolverU}{39}
\newcommand{\resIsoMainBelowTransolverP}{37}
\newcommand{\resIsoMainBelowDilU}{43}

\newcommand{\resIsoSmallU}{7.67}
\newcommand{\resIsoSmallP}{11.49}

\newcommand{\resIsoTinyU}{9.43}
\newcommand{\resIsoTinyP}{14.35}

\newcommand{\resIsoTinyParams}{1.68}

\newcommand{\resIsoXU}{8.54}
\newcommand{\resIsoXP}{14.02}

\newcommand{\resIsoHoursMin}{1.2}
\newcommand{\resIsoHoursMax}{3.5}
\newcommand{\resIsoMainStageOneEpochs}{32}

\newcommand{\resIsoMainStageOneLastGainPct}{0.9}
\newcommand{\resIsoXStageOneEpochs}{23}
\newcommand{\resIsoXSmallStageOneEpochs}{28}

\newcommand{\resMhdOneStepMin}{0.140}
\newcommand{\resMhdOneStepMax}{0.177}
\newcommand{\resMhdWinAMin}{0.689}
\newcommand{\resMhdWinAMax}{0.720}
\newcommand{\resMhdWinBMin}{1.026}
\newcommand{\resMhdWinBMax}{1.041}
\newcommand{\resMhdVsWalrusEarlyMin}{5}
\newcommand{\resMhdVsWalrusEarlyMax}{9}
\newcommand{\resMhdVsWalrusLateMin}{6}
\newcommand{\resMhdVsWalrusLateMax}{10}
\newcommand{\resMhdMaxSdWin}{0.011}

\newcommand{\resMhdHoursAllMin}{1.1}
\newcommand{\resMhdHoursAllMax}{3.5}
\newcommand{\resMhdBlocksMin}{1}
\newcommand{\resMhdBlocksMax}{3}
\newcommand{\resMhdStageOneEpochsAllMin}{3}
\newcommand{\resMhdStageOneEpochsAllMax}{16}
\newcommand{\resBenchHoursMax}{3.5}

\newcommand{\resMhdBelowFnoWinAMin}{42}
\newcommand{\resMhdBelowFnoWinAMax}{44}
\newcommand{\resMhdBelowFnoWinBMin}{35}
\newcommand{\resMhdBelowFnoWinBMax}{36}
\newcommand{\resMhdBelowRetrainWinAMin}{39}
\newcommand{\resMhdBelowRetrainWinAMax}{42}
\newcommand{\resMhdBelowRetrainWinBMin}{44}
\newcommand{\resMhdBelowRetrainWinBMax}{45}
\newcommand{\resMhdStepLastMin}{1.22}
\newcommand{\resMhdStepLastMax}{1.28}
\newcommand{\resMhdStepAboveOneMin}{18}
\newcommand{\resMhdStepAboveOneMax}{19}
\newcommand{\resMhdWalrusRatioMin}{20}
\newcommand{\resMhdWalrusRatioMax}{62}

\newcommand{\resIsoUnetMaxSd}{0.06}

\newcommand{\resIsoAblOneStepDeltaU}{13}

\newcommand{\resIsoAblScratchDeltaU}{17}

\newcommand{\resIsoAblScratchUpdates}{13,200}
\newcommand{\resIsoAblNullMaxAbsDeltaU}{2.3}
\newcommand{\resIsoAblMaxU}{8.97}
\newcommand{\resIsoAblMaxP}{13.53}

\newcommand{\pubIsoUnetU}{35.76}
\newcommand{\pubIsoUnetP}{48.49}

\newcommand{\pubMhdEddyformerOneStep}{0.1160}
\newcommand{\pubMhdWalrusOneStep}{0.0580}

\newcommand{\pubMhdWalrusWalrusLate}{1.2256}

\newcommand{\resCnextRetrainStepLast}{2.05}

\newcommand{\resCnextRetrainStepAboveOne}{8}

\newcommand{\resMSplit}{test}
\newcommand{\resMSims}{50}

\newcommand{\resMTauEnd}{1}
\newcommand{\resMTauMid}{0.48}
\newcommand{\resMTauOne}{0.04}

\newcommand{\resMReOneKXfU}{0.018}

\newcommand{\resMReOneKXfNoiseMaxPct}{103}

\newcommand{\resMReThreeKUnetUOne}{0.002}

\newcommand{\resMReThreeKUnetU}{0.029}

\newcommand{\resMReThreeKUnetNoiseMaxPct}{3}
\newcommand{\resMReThreeKUnetStageOneHours}{20}
\newcommand{\resMReThreeKUnetStageOneU}{0.048}
\newcommand{\resMReThreeKUnetRolloutGainUPct}{47}

\newcommand{\resMReThreeKUnetStageOneJ}{0.172}
\newcommand{\resMReThreeKUnetRolloutGainJPct}{44}
\newcommand{\resMReThreeKUnetRolloutUpdateCost}{4.6}

\newcommand{\resMReThreeKXfU}{0.049}

\newcommand{\resMReThreeKXfJ}{0.182}

\newcommand{\resMReThreeKXfHours}{21}

\newcommand{\resMReThreeKXfStopBlockRiseMinPct}{9}
\newcommand{\resMReThreeKXfStopBlockRiseMaxPct}{15}

\newcommand{\resMReThreeKXfNoiseMaxPct}{2}
\newcommand{\resMReThreeKXfStageOneHours}{11}
\newcommand{\resMReThreeKXfStageOneU}{0.126}

\newcommand{\resMReThreeKXfStageOneJ}{0.406}

\newcommand{\resMReThreeKFnoUOne}{0.020}

\newcommand{\resMReSixKUnetU}{0.053}

\newcommand{\resMReSixKUnetHours}{65}

\newcommand{\resMReSixKUnetEnstPct}{99}
\newcommand{\resMReSixKUnetJsqPct}{97}
\newcommand{\resMReSixKUnetNoiseMaxPct}{1}

\newcommand{\resMReSixKUnetRolloutUpdateCost}{4.6}

\newcommand{\resMReSixKUnetSpecUNThirtyTwo}{0.96}

\newcommand{\resMReSixKUnetSpecBNFortyEight}{0.97}

\newcommand{\resMReSixKXfHours}{36}

\newcommand{\resMReSixKXfEnstPct}{99}
\newcommand{\resMReSixKXfJsqPct}{97}

\newcommand{\resMReSixKXfRolloutUpdateCost}{2.8}

\newcommand{\resMReSixKXfSpecUNThirtyTwo}{0.97}

\newcommand{\resMReSixKXfSpecBNFortyEight}{0.95}

\newcommand{\resMReSixKFnoEnstPct}{69}
\newcommand{\resMReSixKFnoJsqPct}{46}

\newcommand{\resMReSixKFnoSpecUNSixteen}{0.34}

\newcommand{\resMReOneKXfBelowUnetMinPct}{46}
\newcommand{\resMReOneKXfBelowUnetMaxPct}{62}

\newcommand{\resMReOneKRelStart}{17}
\newcommand{\resMReOneKRelEnd}{9}
\newcommand{\resMReOneKKmaxEtaMin}{4.94}
\newcommand{\resMReOneKEkinEndPct}{31}

\newcommand{\resMReOneKTurnovers}{0.9}

\newcommand{\resMReThreeKUnetStageOneVsXfMaxPct}{10}

\newcommand{\resMReThreeKUnetBelowXfMinPct}{41}
\newcommand{\resMReThreeKUnetBelowXfMaxPct}{42}

\newcommand{\resMReThreeKRelStart}{50}
\newcommand{\resMReThreeKRelEnd}{25}
\newcommand{\resMReThreeKKmaxEtaMin}{2.59}
\newcommand{\resMReThreeKEkinEndPct}{54}

\newcommand{\resMReThreeKTurnovers}{1.2}

\newcommand{\resMReSixKSpecUAboveThirtyTwoPct}{0.1}

\newcommand{\resMReSixKSpecBAboveFortyEightPct}{0.6}

\newcommand{\resMReSixKUnetBelowXfMinPct}{5}
\newcommand{\resMReSixKUnetBelowXfMaxPct}{12}

\newcommand{\resMReSixKXfAboveUnetMinPct}{5}
\newcommand{\resMReSixKXfAboveUnetMaxPct}{14}

\newcommand{\resMReSixKRelStart}{101}
\newcommand{\resMReSixKRelEnd}{44}
\newcommand{\resMReSixKKmaxEtaMin}{1.63}
\newcommand{\resMReSixKEkinEndPct}{66}

\newcommand{\resMReSixKTurnovers}{1.3}

\newcommand{\resMReOneKSolverSecPerStep}{5.9}

\newcommand{\resMReOneKSolverCpuHoursPerSim}{52}

\newcommand{\resMReThreeKSolverSecPerStep}{5.7}
\newcommand{\resMReThreeKSolverRanks}{32}
\newcommand{\resMReThreeKSolverWallHoursPerSim}{1.6}
\newcommand{\resMReThreeKSolverCpuHoursPerSim}{50}

\newcommand{\resMReSixKSolverSecPerStep}{5.6}

\newcommand{\resMReSixKSolverCpuHoursPerSim}{49}

\newcommand{\resMRecipeTwoStageHours}{41}

\newcommand{\resMRecipeScratchOverU}{1.3}

\newcommand{\resMRecipeScratchOverB}{1.5}

\newcommand{\resMRecipeScratchOverW}{1.5}

\newcommand{\resMRecipeScratchOverJ}{2.2}

\renewcommand{\algorithmicrequire}{\textbf{Input:}}
\renewcommand{\algorithmicensure}{\textbf{Output:}}
\renewcommand{\algorithmiccomment}[1]{\hfill$\triangleright$~#1}

\graphicspath{{figures/}}

\icmltitlerunning{Learned Surrogates of 3D MHD Turbulence}

\begin{document}

\twocolumn[
  \icmltitle{Learned Surrogates of Three-Dimensional Magnetohydrodynamic Turbulence}

    \begin{icmlauthorlist}
  \icmlauthor{Semih Kacmaz}{ncsa,physics}
  \icmlauthor{E. A. Huerta}{physics,argonne,uchicago,astronomy}
\end{icmlauthorlist}

\icmlaffiliation{ncsa}{National Center for Supercomputing Applications, University of Illinois Urbana-Champaign, Urbana, Illinois, USA}
\icmlaffiliation{physics}{Department of Physics, University of Illinois Urbana-Champaign, Urbana, Illinois, USA}
\icmlaffiliation{argonne}{Data Science and Learning Division, Argonne National Laboratory, Lemont, Illinois, USA}
\icmlaffiliation{uchicago}{Department of Computer Science, The University of Chicago, Chicago, Illinois, USA}
\icmlaffiliation{astronomy}{Department of Astronomy, University of Illinois Urbana-Champaign, Urbana, Illinois, USA}

\icmlcorrespondingauthor{Semih Kacmaz}{skacmaz2@illinois.edu}

  \icmlkeywords{Machine Learning, Scientific Machine Learning, Turbulence, Magnetohydrodynamics, Surrogate Modeling}

  \vskip 0.3in
]

\printAffiliationsAndNotice{}

\begin{abstract}
Three-dimensional magnetohydrodynamic (MHD) turbulence, the flow of astrophysical and fusion plasmas, is costly to
simulate. We present, to our knowledge, the first learned surrogates of three-dimensional resistive MHD
turbulence, and compare three standard backbones in that role: a convolutional UNet, a shifted-window transformer and a
Fourier neural operator. They are trained as autoregressive emulators of decaying MHD turbulence at a resolution of
$\grid{128}$ and Reynolds numbers of 1000, 3000 and 6000, on 500 simulations per Reynolds number, each with one two-stage
procedure, single steps and then short rollouts, and stopped by one rule. Over a window of about one
large-eddy turnover time, the UNet and the transformer reach relative velocity errors of \resMReOneKXfU{} to
\resMReSixKUnetU{} and keep the current sheets, the enstrophy, the current density and the energy spectra of the
simulation; which of the two is the more accurate depends on the Reynolds number, the UNet leading at 3000 and 6000 and
the transformer at 1000. Applied with the same backbones and training procedure to two public benchmarks, the models reach the lowest error reported to date on
the isotropic turbulence set of FactFormer, \resIsoMainBelowTransolverU\,\% below the best published model. On the
compressible MHD set of The Well, their rollout errors are a third to a half below those of the best of The Well's
baselines. They also match a 1.3\,B-parameter pretrained foundation model in its rollout windows, with
\resMhdWalrusRatioMin{}--\resMhdWalrusRatioMax{} times fewer parameters. The code, the configuration of every model, the evaluation scripts and the data generator with the seed of
every simulation will be released upon publication.
\end{abstract}

\section{Introduction}
\label{sec:intro}

\begin{figure*}[t]
  \centering
  \includegraphics{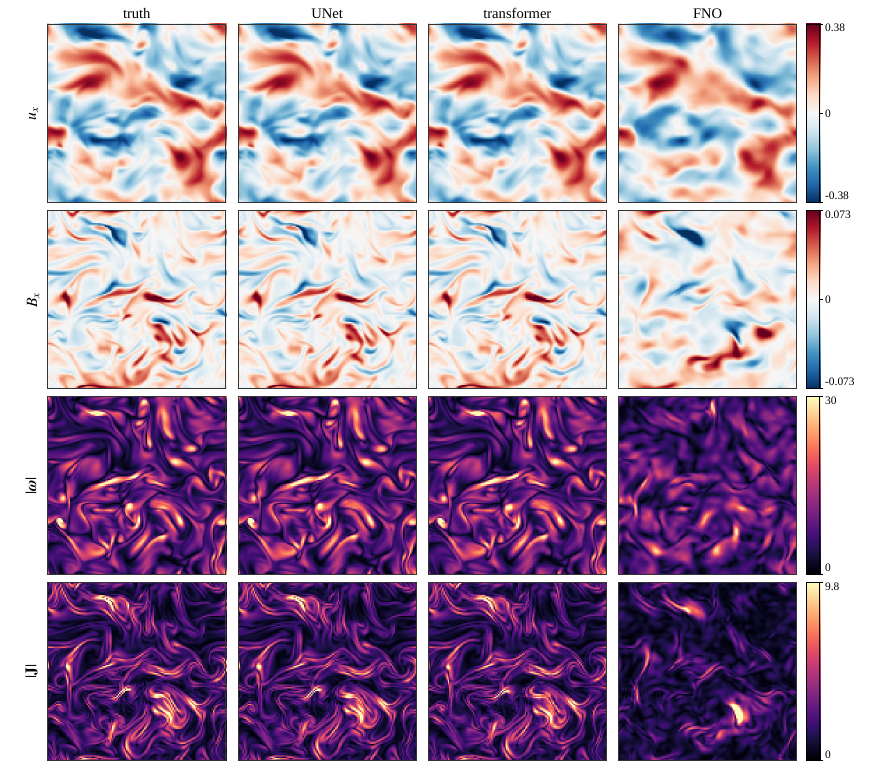}
  \caption{\textbf{Three-dimensional turbulent MHD at $\Re = 6000$ after a rollout of about one eddy-turnover time,
  predicted by three backbones trained with one procedure.} Mid-plane slices ($z = L/2$) of the velocity $u_x$, the
  magnetic field $B_x$, the vorticity magnitude $|\boldsymbol{\omega}|$ and the current density $|\mathbf{J}|$ in the
  $\grid{128}$ volume after 25 autoregressive steps from the initial frame ($\tau = \resMTauEnd$), for the truth and for
  the UNet, the transformer and the FNO. Each row shares the truth's colour scale. Errors in \cref{tab:mhd128}, other
  Reynolds numbers in App.~\ref{app:results}.}
  \label{fig:mhd128_slices}
\end{figure*}

Three-dimensional magnetohydrodynamic (MHD) turbulence, the turbulent motion of an electrically conducting fluid coupled
to a magnetic field, pervades astrophysical plasmas \citep{biskamp2003mhd,goedbloed2004principles,schekochihin2022mhd}. It shapes the solar wind
and the interstellar gas in which stars form \citep{cho2003compressible,burkhart2009density},
and nonlinear MHD governs the edge of tokamak fusion plasmas \citep{gopakumar2024plasma}. Few of these flows can be measured directly
\citep{schekochihin2022mhd}, so numerical simulation is the main way to study them. A simulation must resolve every scale
from the largest eddies down to the viscous and resistive scales, where the magnetic field is folded into thin current
sheets \citep{schekochihin2022mhd}, and its cost grows steeply with the Reynolds number \citep[Ch.~9]{pope2000turbulent}.
A learned surrogate is a network, trained on such simulations, that advances the full three-dimensional state step by
step and runs in their place at a fraction of their cost. It would make feasible studies that need many simulations
and are ruled out today by the solver's cost, such as parameter surveys, design and control iterations \citep{gopakumar2024plasma}, and inference from
observations \citep{ohana2024well}. \Cref{fig:mhd128_slices} shows what the surrogates of this paper achieve at their hardest setting. After 25
autoregressive steps through a window of about one large-eddy turnover time at $\Re = 6000$, a convolutional UNet and a
shifted-window transformer still reproduce the velocity, the magnetic field and the current sheets of a $\grid{128}$
simulation.

Such a surrogate is an autoregressive model in which a network predicts the next state from the current one and a
rollout feeds each prediction back as the next input. Errors compound along a rollout, and two remedies for this are established:
noise on the training inputs \citep{sanchezgonzalez2020learning,stachenfeld2022coarse} and training the model on its own unrolled
predictions \citep{um2020solver,brandstetter2022mppde}. Both are part of our training procedure. The choice of
network is less settled. Much recent work designs transformers for large grids, three-dimensional ones among them
\citep{li2023factformer,p3d2025,du2025eddyformer,transolversigma2026}, or pretrains models across physical systems
\citep{mccabe2023mpp,herde2024poseidon,mccabe2025walrus}. In two comparisons on three-dimensional turbulence, however, a
convolutional network was the most accurate model \citep{li2023factformer} or level with the specialized designs in
one-step error \citep{scalesplit2026}. Learned surrogates of incompressible MHD turbulence have been two-dimensional
\citep{kacmaz2025dinos}. In three dimensions, the main source of MHD turbulence for learned models has
been one public data set of compressible MHD at a resolution of $\grid{64}$, on which general architectures and
foundation models are evaluated \citep{ohana2024well,du2025eddyformer,mccabe2025walrus}.

This paper compares standard backbones, the networks inside such a surrogate, on three-dimensional MHD turbulence under
one training procedure. We generate three data sets of decaying, incompressible, resistive MHD turbulence at a resolution
of $\grid{128}$ and Reynolds numbers $\Re = 1000$, $3000$ and $6000$, with 500 simulations each and Taylor-microscale
Reynolds numbers $\Rel$ up to \resMReSixKRelStart{} (\cref{sec:mhd128:data}). On them we train a convolutional UNet
\citep{ronneberger2015unet,cicek2016unet3d}, a shifted-window transformer \citep{liu2021swin} and a Fourier neural
operator \citep[FNO;][]{li2021fno}. Each network predicts the change of the state from the two most recent frames
\citep{price2023gencast}. It is trained first on single steps with noise on its inputs and then on short rollouts of its own
predictions \citep{um2020solver,oas2026}, and its training is stopped by one rule, the same for every backbone and Reynolds number
(\cref{sec:method}). We evaluate the rollouts over a window of about one large-eddy turnover time on the velocity, the
magnetic field, the vorticity, the current density and the energy spectra (\cref{sec:mhd128}). We then apply the same backbones and two-stage training, adapted to each benchmark's fields, grid and input length, to two
public benchmarks of three-dimensional turbulence
(\cref{sec:benchmarks}). Our contributions are the following.
\begin{itemize}
  \setlength{\itemsep}{2pt}
  \item \textbf{The first learned surrogates of three-dimensional resistive MHD turbulence, to our knowledge, with a
  regime-dependent ranking of two standard backbones.} Trained with one procedure, the UNet and the transformer follow
  the flow at $\grid{128}$ for about one large-eddy turnover time. The relative $L_2$ error of the velocity for the
  better of the two is \resMReOneKXfU{} at $\Re = 1000$ and \resMReSixKUnetU{} at $\Re = 6000$ (\cref{tab:mhd128}).
  Their predictions keep the small scales. The current sheets stay in place, and the enstrophy (the mean squared
  vorticity) and the mean squared current density stay within a few per cent of the truth's. The energy spectra follow
  the truth's up to the wavenumbers beyond which it holds only \resMReSixKSpecUAboveThirtyTwoPct\,\% of its kinetic
  energy at $\Re = 6000$. Which backbone is the more
  accurate depends on the Reynolds number: the UNet leads at $\Re = 3000$ by
  \resMReThreeKUnetBelowXfMinPct{}--\resMReThreeKUnetBelowXfMaxPct\,\% and at $\Re = 6000$ by
  \resMReSixKUnetBelowXfMinPct{}--\resMReSixKUnetBelowXfMaxPct\,\%, where the transformer trains in \resMReSixKXfHours{}
  GPU-hours against the UNet's \resMReSixKUnetHours{}; the transformer leads at $\Re = 1000$ by
  \resMReOneKXfBelowUnetMinPct{}--\resMReOneKXfBelowUnetMaxPct\,\%. The FNO trails both at every Reynolds number.
  \item \textbf{A controlled comparison.} The three backbones are trained on the same simulations with one procedure
  and one stopping rule, fixed in advance and the same for every backbone and Reynolds number. Every choice is made on
  validation simulations, and the test split is scored once, with the final models. The benchmark rows follow protocols
  written before the runs and include models matched in size to the published ones.
  \item \textbf{A data generator and the seed of every simulation}, from which the three $\grid{128}$ data sets can be regenerated,
  to be released upon publication with the code, the configuration of every model and the evaluation scripts
  (App.~\ref{app:repro}).
  \item \textbf{Transfer to public benchmarks, and evidence for the training procedure.} On the isotropic turbulence
  benchmark of FactFormer \citep{li2023factformer}, the UNet reaches the lowest error reported to date,
  \resIsoMainBelowTransolverU\,\% below the best published model \citep{transolversigma2026}, and remains ahead of every published model at that
  model's parameter count. The MHD data set of The Well \citep{ohana2024well} holds forced, compressible and strongly
  magnetized turbulence. On it our models reach rollout errors a third to a half below the best of The Well's baselines,
  and comparable to those of Walrus \citep{mccabe2025walrus}, a 1.3\,B-parameter foundation model pretrained on 19 data
  sets including this one. Our models have \resMhdWalrusRatioMin{}--\resMhdWalrusRatioMax{} times fewer parameters.
  One-step training followed by rollout training beats rollout training from random initialization at equal GPU time,
  by a factor of \resMRecipeScratchOverU{} in velocity and \resMRecipeScratchOverJ{} in current density at
  $\Re = 3000$ (App.~\ref{app:training}). On the isotropic benchmark, where we vary one element of the UNet at a time,
  the two training stages matter more than any element of the network (\cref{sec:bench:iso}).
\end{itemize}

\section{Related Work}
\label{sec:related}

\paragraph{Neural surrogates for turbulence.}
Dilated ResNets \citep{stachenfeld2022coarse} and modern UNets \citep{gupta2022pdearena}, convolutional networks that act locally at several
scales, are strong baselines for fluid flows. Much recent work designs specialized transformers
for large grids and meshes, with factorized attention \citep{li2023factformer}, attention over learned physical
slices \citep{wu2024transolver,transolversigma2026}, shifted windows
\citep{holzschuh2025pdetransformer,pest2026}, hybrid convolution--attention backbones
\citep{p3d2025} or spectral-element tokens \citep{du2025eddyformer}. Local convolutional models nevertheless remain competitive in 3D. In the comparison that introduced the isotropic turbulence benchmark of
FactFormer, the most accurate model was a dilated ResNet \citep{li2023factformer}. In a recent comparison on one regime
of a public compressible 3D MHD data set, a UNet matched P3D and EddyFormer in one-step error \citep{scalesplit2026}. Our results extend this evidence. Trained with the same
procedure, a convolutional UNet and a shifted-window transformer, both local multi-scale backbones, emulate 3D MHD turbulence
accurately at every Reynolds number we study, and on that isotropic benchmark the same approach improves on the best
published models (\cref{sec:benchmarks}).

\paragraph{Stable autoregressive rollouts.}
A one-step model that is fed its own predictions leaves the distribution it was trained on, and its errors compound.
One remedy perturbs the training inputs with noise \citep{sanchezgonzalez2020learning,stachenfeld2022coarse}. Another trains on the model's own predictions, either
without back-propagating through them \citep[the pushforward trick;][]{brandstetter2022mppde} or by unrolling several steps with the
gradient taken through the rollout \citep{um2020solver,koehler2024apebench}. Others refine each prediction with a few denoising or flow-matching steps
\citep{lippe2023pderefiner,flowrefiner2026}, or train at perturbed states labelled by the numerical solver \citep{oas2026}.
Among these remedies, unrolled training gave the lowest short-horizon error in a controlled comparison on isotropic
turbulence at a resolution of $\grid{128}$ \citep{oas2026}. Our models combine unrolled training with noise on their
inputs, and need neither refinement passes nor solver calls. Generative models instead learn a distribution over
future states \citep{kohl2023acdm,price2023gencast} or sample turbulent states directly, without a rollout \citep{lienen2024zero}; we keep a single
deterministic rollout and discuss ensembles and generative models as a next step in \cref{sec:discussion}.

\paragraph{Foundation models and task-specific models.}
Pretraining across physical systems, begun with multiple-physics pretraining \citep{mccabe2023mpp}, has produced
foundation models for PDE emulation \citep{hao2024dpot,herde2024poseidon}. The most recent, Walrus \citep{mccabe2025walrus}, is a 1.3B-parameter
transformer pretrained on 19 data sets, including the 3D MHD turbulence of The Well, a large public collection of
physics simulations \citep{ohana2024well}. Concurrent work argues that scale is a poor
substitute for architectural inductive bias, since models of a few million parameters compete with foundation models on eight 2D
data sets of The Well, second only to Walrus \citep{wavelit2026}. We extend this observation to 3D MHD turbulence. On this MHD data set, task-specific
models with a few tens of millions of parameters, trained from scratch, reach rollout errors comparable to those of Walrus
(\cref{sec:benchmarks}).

\paragraph{MHD surrogates.}
Earlier learned surrogates for MHD turbulence are two-dimensional, such as physics-informed neural operators for incompressible MHD
\citep{rosofsky2023pino} and their combination with diffusion models for MHD turbulence \citep{kacmaz2025dinos}. A related
line models the edge plasma of tokamaks \citep{gopakumar2024plasma}. In 3D, learned models of MHD turbulence have been
evaluated mainly on The Well's MHD data set (MHD\_64), forced, compressible turbulence at a resolution of $\grid{64}$
\citep{ohana2024well,du2025eddyformer,scalesplit2026,mccabe2025walrus}. We complement it with incompressible, decaying MHD
turbulence at a resolution of $\grid{128}$ and three Reynolds numbers. In such flows the current density concentrates in thin sheets
\citep{schekochihin2022mhd}. As found for hydrodynamic turbulence, a low pointwise error can hide distorted small
scales \citep{tide2026}, so we
evaluate vorticity, current density and energy spectra alongside the field itself.

\section{Method}
\label{sec:method}

\begin{figure*}[t]
  \centering
  \includegraphics[width=\textwidth]{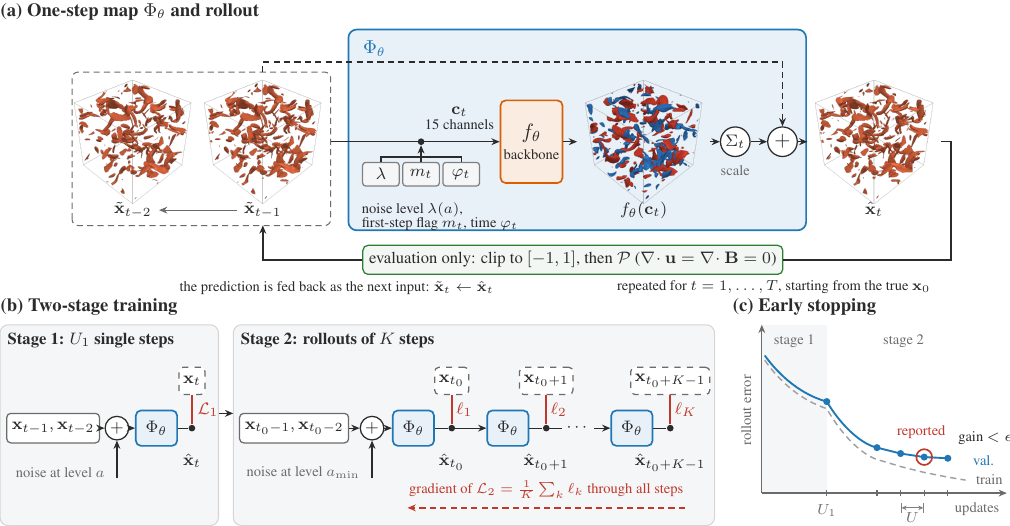}
  \caption{\textbf{Method.} \emph{(a)}~The one-step map $\Phi_\theta$ (blue panel; \cref{eq:onestep}). $\mathbf{x}_t$ is the
  true frame, $\hat{\mathbf{x}}_t$ the prediction and $\tilde{\mathbf{x}}_t$ the frame as the network receives it, noisy
  in training and the previous prediction in a rollout. The two received frames and three scalars, the noise level
  $\lambda$, the first-step flag and the time, form the input $\mathbf{c}_t$ of the backbone $f_\theta$, which predicts
  the change; $\Sigma_t$ scales it per component and step, and the most recent frame is added back. A rollout feeds every
  prediction back as the next input, $T$ times from $\mathbf{x}_0$, and the previous input moves to the second slot (grey
  arrow). At evaluation the prediction is first clipped and projected onto divergence-free fields (\cref{eq:rollout});
  training rollouts skip this. The volumes are isosurfaces of the vorticity magnitude of a training simulation at
  $\Re = 3000$ and of the signed change of one velocity component (red positive, blue negative). \emph{(b)}~The two
  training stages as computation graphs (dashed boxes: true frames; dots: predictions; red links: losses): single steps
  from noisy true frames, then rollouts of $K$ steps with a loss at every step and the gradient through the whole window.
  Each $\Phi_\theta$ receives the two most recent frames; the chain draws the latest. \emph{(c)}~Early stopping of the
  second stage on the validation rollout error, checked every $U$ updates; the model before the first check that gains
  less than $\epsilon$ is reported. Symbols and values in \cref{tab:settings}.}
  \label{fig:method}
\end{figure*}

We emulate the flow with a deterministic network that advances the full three-dimensional state by one time step and is
applied repeatedly to its own predictions. The network is trained in two stages, first on single steps and then on short
rollouts of its own predictions. Three backbones share this interface and this training: a convolutional UNet, a
shifted-window transformer and a Fourier neural operator (FNO). \Cref{fig:method} summarizes the approach.

\subsection{Problem setting}
\label{sec:method:problem}

We consider decaying turbulence governed by the incompressible resistive MHD equations \citetext{\citealp{biskamp2003mhd}; \citealp[Ch.~4]{goedbloed2004principles}; \citealp[Ch.~14]{goedbloed2010advanced}} in the
periodic unit box. With unit density and the magnetic field in velocity units, the velocity $\mathbf{u}$ and the magnetic
field $\mathbf{B}$ obey
\begin{equation}
  \begin{aligned}
    \partial_\tau \mathbf{u} &= -(\mathbf{u}\cdot\nabla)\mathbf{u} + (\mathbf{B}\cdot\nabla)\mathbf{B} - \nabla p + \nu \nabla^2 \mathbf{u}, \\
    \partial_\tau \mathbf{B} &= \nabla\times(\mathbf{u}\times\mathbf{B}) + \eta \nabla^2 \mathbf{B},
  \end{aligned}
  \label{eq:mhd}
\end{equation}
with $\nabla\cdot\mathbf{u} = \nabla\cdot\mathbf{B} = 0$, where $p$ is the total pressure, $\nu = 1/\Re$ the viscosity and
$\eta = 1/\Re_M$ the magnetic diffusivity. No forcing acts, so the turbulence decays. All data sets use $\Re_M = \Re$,
so the magnetic Prandtl number $\mathrm{Pr}_M = \nu/\eta$ is 1 (\cref{sec:mhd128} gives the Reynolds numbers, initial
conditions and solver). A simulation provides frames $\mathbf{s}_t \in \mathbb{R}^{6 \times N^3}$,
the components of $\mathbf{u}$ and $\mathbf{B}$ at time $\tau = t\,\Delta t$ on a grid of $N^3 = \grid{128}$ points, for
$t = 0, \dots, T$ with $T = 25$. \Cref{eq:mhd} is autonomous and first order in time, and $p$ follows from $\mathbf{u}$ and
$\mathbf{B}$, so one frame determines the next: $\mathbf{s}_t = \mathcal{S}(\mathbf{s}_{t-1})$, with $\mathcal{S}$ the flow
of \cref{eq:mhd} over $\Delta t$.

Given only $\mathbf{s}_0$ of an unseen simulation, we predict $\mathbf{s}_1, \dots, \mathbf{s}_T$ by an
\emph{autoregressive rollout}: a learned one-step map approximates $\mathcal{S}$ and is applied $T$ times, each time to its
own previous outputs (\cref{fig:method}a). Besides the fields we assess the vorticity $\boldsymbol{\omega} =
\nabla\times\mathbf{u}$ and the current density $\mathbf{J} = \nabla\times\mathbf{B}$, which emphasize the small scales
where energy is dissipated.

\subsection{One-step model}
\label{sec:method:model}

Each component is scaled to $[-1, 1]$ by its minimum and maximum over the training simulations; $\mathcal{N}$ denotes this
map and $\mathbf{x}_t = \mathcal{N}(\mathbf{s}_t)$ the scaled frames. A tilde marks a frame as the network receives it,
noisy in training or predicted in a rollout. The backbone network $f_\theta$, one of the three networks of
\cref{sec:method:backbones}, receives the concatenation of 15 channels
\begin{equation}
  \mathbf{c}_t = \bigl[\tilde{\mathbf{x}}_{t-1},\, \tilde{\mathbf{x}}_{t-2},\, \lambda,\, m_t,\, \varphi_t\bigr],
  \label{eq:input}
\end{equation}
where three scalars are broadcast to the grid: the input noise level $\lambda$ (\cref{sec:method:training}), a first-step
flag $m_t$ and the time $\varphi_t = 2t/T - 1$. Since one frame determines the next, $\tilde{\mathbf{x}}_{t-1}$ would
suffice in principle. The second frame makes the time derivative available directly, as the difference
$(\tilde{\mathbf{x}}_{t-1} - \tilde{\mathbf{x}}_{t-2})/\Delta t$; from one frame, the network would have to evaluate the
right-hand side of \cref{eq:mhd}, including the non-local pressure. At $t = 1$ no earlier frame exists, so its slot holds
zeros and $m_1 = -1$ ($m_t = 1$ otherwise); a copy of $\mathbf{x}_0$ there degraded the first step in a development
test (App.~\ref{app:training}).

The prediction $\hat{\mathbf{x}}_t$ adds the scaled change predicted by the network to the most recent frame,
\begin{equation}
  \hat{\mathbf{x}}_t = \Phi_\theta(\tilde{\mathbf{x}}_{t-1}, \tilde{\mathbf{x}}_{t-2})
  := \tilde{\mathbf{x}}_{t-1} + \Sigma_t\, f_\theta(\mathbf{c}_t),
  \label{eq:onestep}
\end{equation}
where the diagonal matrix $\Sigma_t$ holds, for each component, the standard deviation of the true change
$\mathbf{x}_t - \mathbf{x}_{t-1}$ over the training simulations. Conditioning on two past states and predicting the normalized change from the most
recent one follows GenCast \citep{price2023gencast}, which uses one scale per variable and level. Our scales also depend
on $t$, because the typical change differs between $\mathbf{u}$ and $\mathbf{B}$ and varies along the decay, by up to a
factor of 3.6 for one component and Reynolds number; the target then has unit variance for every component and step. Since $\Sigma_t$ depends
on $t$, the network needs $\varphi_t$ even though \cref{eq:mhd} is autonomous.

Both training stages regress $f_\theta(\mathbf{c}_t)$ onto the scaled change from the frame \emph{as received},
$\Sigma_t^{-1}(\mathbf{x}_t - \tilde{\mathbf{x}}_{t-1})$. With $\lVert\cdot\rVert^2$ the mean square over the $6N^3$
entries, \cref{eq:onestep} gives
\begin{equation}
  \bigl\lVert f_\theta(\mathbf{c}_t) - \Sigma_t^{-1}(\mathbf{x}_t - \tilde{\mathbf{x}}_{t-1}) \bigr\rVert^2
  = \bigl\lVert \Sigma_t^{-1}(\hat{\mathbf{x}}_t - \mathbf{x}_t) \bigr\rVert^2 .
  \label{eq:identity}
\end{equation}
Learning the change is thus learning the next frame, weighted per component and step. Because the change is measured from
the received frame, the target also removes the input's error $\tilde{\mathbf{x}}_{t-1} - \mathbf{x}_{t-1}$.

\subsection{Rollout}
\label{sec:method:rollout}

A rollout starts from $\tilde{\mathbf{x}}_0 = \mathbf{x}_0$ and applies \cref{eq:onestep} for $t = 1, \dots, T$ to the
two most recent received frames (at $t = 1$ the second slot is empty, \cref{sec:method:model}). Each prediction becomes
the next input after it is clipped to the training range $[-1, 1]$ and projected onto divergence-free fields,
\begin{equation}
  \tilde{\mathbf{x}}_t = \mathcal{N}\bigl(\mathcal{P}\,\mathcal{N}^{-1}(\operatorname{clip}\hat{\mathbf{x}}_t)\bigr),
  \label{eq:rollout}
\end{equation}
without added noise ($\lambda$ at its lowest training value). The projection $\mathcal{P}$ acts on $\mathbf{u}$ and
$\mathbf{B}$ separately, in physical units, because $\mathcal{N}$ scales the components of a field differently. In Fourier
space it removes the component of each mode along its wavevector $\mathbf{k} \neq \mathbf{0}$,
\begin{equation}
  \widehat{\mathcal{P}\mathbf{v}}(\mathbf{k}) = \bigl(\mathbf{I} - \mathbf{k}\mathbf{k}^{\top} / \lvert\mathbf{k}\rvert^2\bigr)\, \hat{\mathbf{v}}(\mathbf{k}),
  \label{eq:leray}
\end{equation}
which is the gradient part of $\mathbf{v}$. Neither clipping nor projection can increase the error of the frame it acts on. $\mathcal{P}$ is an orthogonal
projection and the true fields are divergence-free, so $\lVert\mathcal{P}\mathbf{v} - \mathbf{u}_t\rVert = \lVert\mathcal{P}(\mathbf{v} - \mathbf{u}_t)\rVert
\leq \lVert\mathbf{v} - \mathbf{u}_t\rVert$ for any prediction $\mathbf{v}$, and likewise for $\mathbf{B}$;
clipping cannot increase the error where the true value lies in $[-1, 1]$. The guarantee holds for that frame only, since both operations change the
inputs of later steps. Training rollouts, by contrast, omit them, so that the loss penalizes
divergence and out-of-range values in the network's raw output.

\subsection{Backbones}
\label{sec:method:backbones}

The three backbones $f_\theta$ map the 15 input channels to six output channels on the full $\grid{128}$ grid; a small search
over width, depth and related settings fixed each configuration (App.~\ref{app:training}). The UNet has 37.7 M parameters,
the transformer 66.3 M and the FNO 37.8 M (complex weights counted as two).

\emph{UNet.} The 3D UNet \citep{ronneberger2015unet,cicek2016unet3d} halves the grid three times, to $\grid{16}$, and
restores it, concatenating the features of matching levels. The levels have 40, 40, 80 and 160 channels, the bottleneck 320,
and two residual blocks each; a block is a $3^3$ convolution, RMS normalization and SiLU, applied twice. One self-attention
layer over all $\grid{16}$ positions of the bottleneck mixes the volume globally. Convolutions pad circularly and the
attention has no positional encoding, so the network commutes exactly with periodic shifts by multiples of 8 voxels. Window
attention at the $\grid{32}$ and $\grid{16}$ levels, tested in the search, was worse at equal cost (App.~\ref{app:training}).

\begin{table}[t]
  \caption{Training settings, shared by all backbones and Reynolds numbers.}
  \label{tab:settings}
  \centering
  \small
  \setlength{\tabcolsep}{3pt}
  \begin{tabular}{@{}l l p{4.6cm}@{}}
    \toprule
    & value & role and choice \\
    \midrule
    $[a_{\min}, a_{\max}]$ & $[0.01, 0.3]$ & input-noise levels; spans the errors of rollout inputs, from early to late steps (App.~B) \\
    $U_1$ & 40,000 & first-stage updates; twice as many improved only $\mathbf{J}$, at 1.4 times the cost (App.~B) \\
    $K$ & 4 & window length, as the unrolled baseline of \citet{oas2026}; memory grows with $K$ \\
    $U$ & 6,600 & updates between early-stopping checks \\
    $\epsilon$ & 5\% & tolerance of the early-stopping rule, set in advance \\
    \bottomrule
  \end{tabular}
\end{table}
\begin{algorithm}[t]
  \caption{Two-stage training with early stopping}
  \label{alg:training}
  \begin{algorithmic}[1]
    \REQUIRE training and validation simulations; settings $a_{\min}, a_{\max}, U_1, K, U, \epsilon$ (\cref{tab:settings})
    \ENSURE weights $\bar{\theta}^{\star}$
    \STATE initialize $\theta$; $\bar{\theta} \gets \theta$
    \FOR{$u = 1$ {\bfseries to} $U_1$}
      \STATE draw $(i, t)$ and $a \sim \mathrm{LogU}[a_{\min}, a_{\max}]$; noisy $\tilde{\mathbf{x}}_{t-1}, \tilde{\mathbf{x}}_{t-2}$
      \STATE $\hat{\mathbf{x}}_t \gets \Phi_\theta(\tilde{\mathbf{x}}_{t-1}, \tilde{\mathbf{x}}_{t-2})$; \textsc{Step}$(\mathcal{L}_1)$ \COMMENT{\cref{eq:loss1}}
    \ENDFOR
    \STATE $E^{(0)} \gets \textsc{Rollouts}(\bar{\theta})$; $b \gets 0$
    \REPEAT
      \STATE $b \gets b + 1$; $\bar{\theta}_{b-1} \gets \bar{\theta}$
      \FOR{$u = 1$ {\bfseries to} $U$ (twice as many if $b = 1$)}
        \STATE draw $(i, t_0)$; noisy true inputs at level $a_{\min}$
        \FOR{$t = t_0$ {\bfseries to} $t_0 + K - 1$}
          \STATE $\hat{\mathbf{x}}_t \gets \Phi_\theta(\tilde{\mathbf{x}}_{t-1}, \tilde{\mathbf{x}}_{t-2})$; $\tilde{\mathbf{x}}_t \gets \hat{\mathbf{x}}_t$
        \ENDFOR
        \STATE \textsc{Step}$(\mathcal{L}_2)$ \COMMENT{\cref{eq:loss2}, through all $K$ steps}
      \ENDFOR
      \STATE $E^{(b)} \gets \textsc{Rollouts}(\bar{\theta})$; $g_b^{\mathrm{val}}, g_b^{\mathrm{tr}}$ \COMMENT{\cref{eq:gain}}
    \UNTIL{$g_b^{\mathrm{val}} < \epsilon$ \OR ($g_b^{\mathrm{tr}} \geq \epsilon$ \AND $g_b^{\mathrm{val}} < g_b^{\mathrm{tr}}/2$)}
    \STATE {\bfseries return} $\bar{\theta}^{\star} = \bar{\theta}_{b-1}$
  \end{algorithmic}
  \vspace{1pt}
  \small \textsc{Step}: gradient on a batch of four, clipped to norm 1; AdamW update; $\bar{\theta} \gets \beta\bar{\theta} +
  (1 - \beta)\theta$. \textsc{Rollouts}: \cref{eq:rollout} on validation and training simulations, errors
  \cref{eq:error} at $t = T$.
\end{algorithm}
\emph{Transformer.} We carry the 2D PDE-Transformer \citep{holzschuh2025pdetransformer} to three dimensions. The volume is split into
tokens of $2^3$ voxels, and a U-shaped hierarchy merges them $2 \times 2 \times 2$ between its four levels, which have 128,
256, 512 and 1024 channels and 2, 2, 4 and 2 transformer blocks from the finest level to the coarsest. In each block,
attention acts within windows of $4^3$ tokens, shifted cyclically between consecutive blocks \citep{liu2021swin}, with
relative positions, and the residual branches are initialized to zero \citep{peebles2023dit}. The output is formed at full
resolution by periodic convolutions over the decoded tokens and the raw input, which keeps patch borders from imprinting
on the smallest scales. The transformer alone is trained with a shift augmentation, in which every window of the second
training stage, inputs and targets together, is shifted by a random periodic translation $\boldsymbol{\delta}$. Because
\cref{eq:mhd} is invariant under such translations, $\mathcal{S}(\mathcal{T}_{\boldsymbol{\delta}}\mathbf{s}) =
\mathcal{T}_{\boldsymbol{\delta}}\mathcal{S}(\mathbf{s})$, a shifted window is an exact sample of the same dynamics.
Inference uses no shift. The augmentation helped the transformer but not the UNet in a development test
(App.~\ref{app:training}).

\emph{FNO.} The Fourier neural operator \citep{li2021fno} lifts the input to 48 channels and applies four layers
$\mathbf{v} \leftarrow \operatorname{GELU}(W\mathbf{v} + \mathcal{F}^{-1} R\, \mathcal{F}\mathbf{v})$, where $R$ acts on
the Fourier modes $-8 \leq n_i < 8$ and discards the rest. Unlike the UNet and the transformer, which resolve fine scales
locally and reach across the box only through their coarse levels, the FNO mixes all positions in every layer through a
few modes.

\subsection{Two-stage training}
\label{sec:method:training}

\paragraph{Stage 1: single steps.}
In the first stage the inputs of every step are true frames, the setting known as teacher forcing. A model later fed its
own predictions meets input errors that clean training frames never show, so, following
\citet{sanchezgonzalez2020learning} and \citet{stachenfeld2022coarse}, we add noise to the true input frames,
$\tilde{\mathbf{x}}_{t-k} = \mathbf{x}_{t-k} + a\,\sigma_{\mathrm{d}}\,\boldsymbol{\epsilon}_k$ for $k = 1, 2$, with
$\boldsymbol{\epsilon}_k \sim \mathcal{N}(\mathbf{0}, \mathbf{I})$ and $\sigma_{\mathrm{d}}$ the standard deviation of the
scaled fields averaged over components. By \cref{eq:identity}, the first stage,
\begin{equation}
  \mathcal{L}_1(\theta) = \mathbb{E}_{i,\,t,\,a,\,\boldsymbol{\epsilon}}
  \bigl\lVert \Sigma_t^{-1}(\hat{\mathbf{x}}_t - \mathbf{x}_t) \bigr\rVert^2 ,
  \label{eq:loss1}
\end{equation}
asks for the clean next frame from a corrupted input, so the network learns to shrink input errors while it advances the
flow. The level $a$ is drawn log-uniformly from $[a_{\min}, a_{\max}]$, a range that covers the relative errors of
rollout inputs from the first steps to late steps (\cref{tab:settings}), and is given to the network as
$\lambda = 2\ln(a/a_{\min}) / \ln(a_{\max}/a_{\min}) - 1$, so that one network learns how far to trust its input at
every level. The expectation runs over training simulations $i$ and steps $t \in \{1, \dots, T\}$; the stage lasts a
fixed budget of $U_1$ updates.

\paragraph{Stage 2: rollouts.}
The second stage continues from the first (weights, optimizer state and weight average) on windows of $K$ steps
(\cref{fig:method}b). A window starts at $t_0 \in \{1, \dots, T - K + 1\}$ from true frames with noise at level
$a_{\min}$; every later input is the model's previous prediction, $\tilde{\mathbf{x}}_t = \hat{\mathbf{x}}_t$, without
noise, clipping or projection. Each step is compared with the true frame,
\begin{equation}
  \mathcal{L}_2(\theta) = \mathbb{E}_{i,\,t_0,\,\boldsymbol{\epsilon}}\, \frac{1}{K} \sum_{t=t_0}^{t_0+K-1}
  \bigl\lVert \Sigma_t^{-1}(\hat{\mathbf{x}}_t - \mathbf{x}_t) \bigr\rVert^2 ,
  \label{eq:loss2}
\end{equation}
and the gradient flows through the whole window, including the base $\tilde{\mathbf{x}}_{t-1} = \hat{\mathbf{x}}_{t-1}$
of each later step. Each step is thus trained on the inputs it meets in a rollout and for its effect on the steps after
it. The labels are the true trajectory, the target of our evaluation.

Training on single steps before unrolled steps is established practice: FourCastNet and GraphCast fine-tune a
single-step model through two and up to twelve unrolled steps \citep{kurth2023fourcastnet,lam2023graphcast}, and
\citet{um2020solver} pre-train with short look-aheads before longer ones. The first stage is also worth its cost. At equal GPU time, about \resMRecipeTwoStageHours{} GPU-hours per UNet
run, rollout training from random initialization ended with \resMRecipeScratchOverU{}, \resMRecipeScratchOverW{} and
\resMRecipeScratchOverJ{} times the two-stage errors in field, vorticity and current density at $t = T$, averaged over
validation simulations and higher in every one of them (App.~\ref{app:training}).

\paragraph{Early stopping.}
The second stage is stopped on the validation rollout error (\cref{fig:method}c). Every $U$ updates (the first time after
$2U$), we roll out the weight-averaged model on validation and training simulations and compute the relative gain
since the previous check,
\begin{equation}
  g_b = \frac{1}{3} \sum_{\mathbf{q} \in \{\mathbf{s}, \boldsymbol{\omega}, \mathbf{J}\}}
  \Bigl(1 - E^{(b)}_{\mathbf{q}} \big/ E^{(b-1)}_{\mathbf{q}}\Bigr),
  \label{eq:gain}
\end{equation}
with $E^{(b)}_{\mathbf{q}}$ the error \cref{eq:error} at $t = T$ after the $b$-th interval ($b = 0$: end of the first
stage). Training stops when the validation gain falls below $\epsilon$, or when the training gain reaches $\epsilon$ and is more
than twice the validation gain (\cref{alg:training}). We report the model of the previous check.

\paragraph{Optimization.}
Both stages use AdamW \citep{loshchilov2019adamw} with four samples (four windows) per update, a linear warm-up to a
constant learning rate, gradients clipped to norm 1, and an exponential moving average of the weights, which is the model
we evaluate (\cref{alg:training}, App.~\ref{app:training}). Rollout training at $\grid{128}$ is memory-bound, so for the UNet the
first step of each window is recomputed in the backward pass and the windows of a batch are back-propagated one at a
time, which leaves the gradient unchanged and fits one window on a GPU.

\subsection{Evaluation}
\label{sec:method:eval}

Every reported number comes from a rollout from the initial frame of a held-out simulation, never re-initialized from the
truth. For $\mathbf{q} \in \{\mathbf{u}, \mathbf{B}, \boldsymbol{\omega}, \mathbf{J}\}$ we report the relative error averaged over
the $M$ simulations of a split,
\begin{equation}
  E_{\mathbf{q}}(t) = \frac{1}{M} \sum_{i=1}^{M}
  \frac{\lVert \hat{\mathbf{q}}^{(i)}_t - \mathbf{q}^{(i)}_t \rVert_2}{\lVert \mathbf{q}^{(i)}_t \rVert_2},
  \label{eq:error}
\end{equation}
over all grid points and components in physical units, with $\boldsymbol{\omega}$ and $\mathbf{J}$ of a prediction
computed spectrally. Energy spectra are shown against the wavenumber $k = 2\pi n$, with $n$ the shell index. Model
choices and stopping use the validation simulations only; the test simulations are evaluated once, with the final models.

\section{Turbulent MHD at $\grid{128}$}
\label{sec:mhd128}
\begin{table*}[t]
  \caption{\textbf{Rollout errors on the $\grid{128}$ MHD data sets.} Relative $L_2$ error (\cref{eq:error}) of velocity,
  magnetic field, vorticity and current density at the end of the window, $\tau = \resMTauEnd$, about one eddy-turnover
  time, on the \resMSplit{} split (\resMSims{} simulations). Lower is better; best per column and Reynolds number in bold.
  Params: trainable parameters. GPU-h: training time of the final model over both stages on GH200 GPUs. $\Rel$:
  Taylor-scale Reynolds number over the stored window (App.~\ref{app:data}).}
  \label{tab:mhd128}
  \centering
  \footnotesize
  \setlength{\tabcolsep}{4pt}
  \begin{tabular}{@{}l r r r r r r@{}}
\toprule
& & & \multicolumn{4}{c@{}}{relative error at $\tau = 1$} \\
\cmidrule(l){4-7}
& Params (M) & GPU-h & $\mathbf{u}$ & $\mathbf{B}$ & $\boldsymbol{\omega}$ & $\mathbf{J}$ \\
\midrule
\multicolumn{7}{@{}l}{\emph{$\Re = 1000$ ($\Rel = 9$--$17$)}} \\
UNet & 37.7 & 20 & 0.048 & 0.068 & 0.069 & 0.094 \\
Transformer & 66.3 & 31 & \textbf{0.018} & \textbf{0.031} & \textbf{0.034} & \textbf{0.051} \\
FNO & 37.8 & 18 & 0.127 & 0.285 & 0.272 & 0.491 \\
\midrule
\multicolumn{7}{@{}l}{\emph{$\Re = 3000$ ($\Rel = 25$--$50$)}} \\
UNet & 37.7 & 65 & \textbf{0.029} & \textbf{0.060} & \textbf{0.062} & \textbf{0.107} \\
Transformer & 66.3 & 21 & 0.049 & 0.101 & 0.107 & 0.182 \\
FNO & 37.8 & 18 & 0.380 & 0.868 & 0.724 & 1.090 \\
\midrule
\multicolumn{7}{@{}l}{\emph{$\Re = 6000$ ($\Rel = 44$--$101$)}} \\
UNet & 37.7 & 65 & \textbf{0.053} & \textbf{0.127} & \textbf{0.133} & \textbf{0.245} \\
Transformer & 66.3 & 36 & 0.055 & 0.134 & 0.152 & 0.266 \\
FNO & 37.8 & 17 & 0.520 & 1.045 & 0.884 & 1.122 \\
\bottomrule
\end{tabular}

\end{table*}

\subsection{Data}
\label{sec:mhd128:data}
We generate three data sets of decaying MHD turbulence from \cref{eq:mhd} at $\Re = \Re_M \in \{1000, 3000, 6000\}$ in a
periodic unit box, with the pseudo-spectral solver Dedalus~3 \citep{burns2020dedalus} on $\grid{128}$ modes, a third-order
implicit--explicit Runge--Kutta scheme with a fixed step of $10^{-3}$, and $3/2$ dealiasing. Each simulation starts from a
divergence-free random field, $\mathbf{u}_0 = \nabla\times\boldsymbol{\Phi}$ and $\mathbf{B}_0 = \nabla\times\mathbf{A}_0$, where every
component of $\boldsymbol{\Phi}$ and $\mathbf{A}_0$ is a periodic Gaussian random field with squared-exponential covariance of
correlation length $0.1$ and standard deviation $0.1$ or $5\times10^{-3}$. The stored
window holds 26 frames of $\mathbf{u}$ and $\mathbf{B}$ at $\Delta t = 0.04$, $\tau \in [0, 1]$. Each Reynolds number has
500 simulations from distinct seeds, split at random into 400 training, 50 validation and 50 test simulations (App.~\ref{app:data}).

The nominal $\Re = 1/\nu$ fixes the viscosity in a unit box. The turbulence itself is characterized by the Taylor-microscale
Reynolds number $\Rel = u'\lambda/\nu$, with $u'$ the root-mean-square velocity component and $\lambda$ the Taylor microscale
\citep[Ch.~6]{pope2000turbulent}, estimated per frame in App.~\ref{app:data}. The flow is unforced and decays over the
window. The frame-mean $\Rel$ falls from \resMReOneKRelStart{} to \resMReOneKRelEnd{} at $\Re = 1000$, from
\resMReThreeKRelStart{} to \resMReThreeKRelEnd{} at $3000$ and from \resMReSixKRelStart{} to \resMReSixKRelEnd{} at $6000$,
and the kinetic energy drops to \resMReOneKEkinEndPct{}, \resMReThreeKEkinEndPct{} and \resMReSixKEkinEndPct{}\,\%
of its initial value. The window spans about one large-eddy turnover time at every $\Re$
(\resMReOneKTurnovers{}--\resMReSixKTurnovers{} turnovers). The magnetic field is
dynamically weak. The flow stretches and folds it, and the thin current sheets this produces (\cref{fig:mhd128_slices}) are
the finest structures a model must follow. The
grid resolves the Kolmogorov scale at every Reynolds number, since the product of the largest resolved wavenumber and the
Kolmogorov scale stays above \resMReSixKKmaxEtaMin{} even at $\Re = 6000$ (App.~\ref{app:data}).

\begin{figure}[t]
  \centering
  \includegraphics[width=\columnwidth]{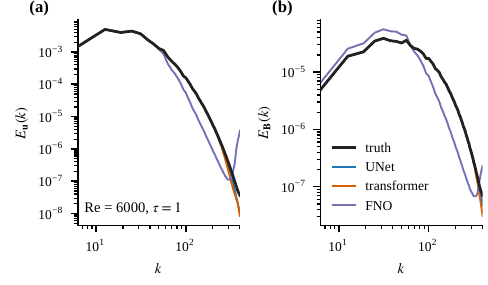}
  \caption{\textbf{Energy spectra at $\tau = \resMTauEnd$, $\Re = 6000$} (\resMSplit{} split, mean over \resMSims{} simulations), against
  the wavenumber $k = 2\pi n$ up to the Nyquist shell: \emph{(a)}~kinetic, \emph{(b)}~magnetic, truth in black.}
  \label{fig:mhd128_spectra}
\end{figure}

\subsection{Setup}
\label{sec:mhd128:setup}
All models are trained as in \cref{sec:method} with the settings of \cref{tab:settings}. The one-step stage runs 40,000 updates. Rollout training
on four-step windows then proceeds in blocks of 13,200 and 6,600 updates, until the stopping rule of \cref{sec:method:training} ends it
(\cref{alg:training}).
The reported model is the one at the end of the last block that passed. Each update uses a batch
of four (frames in the one-step stage, windows in the rollout stage) on one to four GH200 GPUs with the same effective batch.
The learning rate is $4\times10^{-4}$ for the UNet and the transformer and $1.6\times10^{-3}$ for the FNO, the interior optimum of a
learning-rate check on the $\grid{64}$ data at $\Re = 1000$ (App.~\ref{app:training}). Every rollout starts from the initial frame of a held-out
simulation and runs $T = 25$ steps to $\tau = 1$. \cref{tab:mhd128} and
the figures report the \resMSplit{} split (\resMSims{} simulations) with the error \cref{eq:error} of $\mathbf{u}$,
$\mathbf{B}$, $\boldsymbol{\omega}$ and $\mathbf{J}$ at the end of the window. Next to the errors we report the ratios of
predicted to true enstrophy $\langle|\boldsymbol{\omega}|^2\rangle$ and mean squared current density
$\langle|\mathbf{J}|^2\rangle$ at the end of the window, which show whether a prediction keeps the small-scale activity of
the truth (a smoothed field loses both), and the energy spectra. Errors at earlier times, the sensitivity to
input noise and the gap between validation and training simulations are reported in App.~\ref{app:results} and
App.~\ref{app:training}. Training cost is the GPU time of both stages, in-training validation included.

\subsection{Results}
\label{sec:mhd128:results}
Which backbone is the most accurate depends on the Reynolds number (\cref{tab:mhd128}). At $\Re = 3000$ the UNet leads
in every quantity, with errors \resMReThreeKUnetBelowXfMinPct{}--\resMReThreeKUnetBelowXfMaxPct{}\,\% below the
transformer's. At $\Re = 6000$ the UNet leads again, by \resMReSixKUnetBelowXfMinPct{}--\resMReSixKUnetBelowXfMaxPct{}\,\%.
At $\Re = 1000$ the transformer leads, with errors \resMReOneKXfBelowUnetMinPct{}--\resMReOneKXfBelowUnetMaxPct{}\,\% below
the UNet's. The UNet's reported model at $\Re = 1000$ is its one-step-stage model, the stopping rule having ended its
rollout training after the first block (App.~\ref{app:training}). The FNO trails both at every Reynolds number, and at
$\Re = 3000$ its one-step velocity error (\resMReThreeKFnoUOne) is already about ten times the UNet's (\resMReThreeKUnetUOne). The
errors grow over the window, except the transformer's vorticity and current density at $\Re = 1000$, which are higher after
the first step than after the last (App.~\ref{app:results}). The current density is the hardest quantity for every model.

The rollouts keep the small-scale activity. At $\Re = 6000$ the enstrophy and the mean squared current density stay at
\resMReSixKUnetEnstPct{} and \resMReSixKUnetJsqPct{}\,\% of the truth for the UNet and at \resMReSixKXfEnstPct{} and
\resMReSixKXfJsqPct{}\,\% for the transformer, whereas the FNO's fall to \resMReSixKFnoEnstPct{} and
\resMReSixKFnoJsqPct{}\,\% (other Reynolds numbers in App.~\ref{app:results}). The kinetic-energy spectra of the UNet and
the transformer at $\Re = 6000$ (\cref{fig:mhd128_spectra}) stay at \resMReSixKUnetSpecUNThirtyTwo{} and \resMReSixKXfSpecUNThirtyTwo{} of
the truth's up to shell 32, beyond which the truth holds only \resMReSixKSpecUAboveThirtyTwoPct{}\,\% of its kinetic energy,
and lose energy only in those highest shells. Their magnetic spectra stay at \resMReSixKUnetSpecBNFortyEight{} and
\resMReSixKXfSpecBNFortyEight{} up to shell 48, with \resMReSixKSpecBAboveFortyEightPct{}\,\% of the magnetic energy above.
The FNO's spectra fall to \resMReSixKFnoSpecUNSixteen{} of the truth by shell 16 and carry spurious energy at the highest
wavenumbers. Gaussian noise of level $0.01$ added to the received frames at every step of the rollout changes the UNet's
errors by at most \resMReThreeKUnetNoiseMaxPct{}\,\% at $\Re = 3000$ and \resMReSixKUnetNoiseMaxPct{}\,\% at $\Re = 6000$, and
the transformer's by \resMReThreeKXfNoiseMaxPct{}\,\% at both. At $\Re = 1000$ the transformer's errors rise by up to
\resMReOneKXfNoiseMaxPct{}\,\% under the same noise while the UNet's are unchanged (four simulations,
App.~\ref{app:results}).

\Cref{fig:mhd128_slices} shows the mid-plane at $\tau = \resMTauEnd$ for $\Re = 6000$, the hardest case. The UNet and the
transformer reproduce the velocity and the magnetic field down to the thin current sheets of $|\mathbf{J}|$, with the sheets
in place and at the right intensity. The FNO's velocity is a smoothed version of the truth, its magnetic field keeps only
the largest structures, and its vorticity and current density lose the sheets.

\begin{table*}[t]
  \caption{\textbf{Public benchmarks} (test splits; ours: mean of three seeds; lower is better, best in bold; standard
  deviations, all rows and the protocols in App.~\ref{app:protocols}). \emph{(a)}~Isotropic turbulence \citep{li2023factformer}: relative $L_2$
  error (\%) of velocity $\mathbf{u}$ and pressure $p$ over 10 predicted frames from 10 given frames (ours use the last
  four). Small / tiny = at FactFormer's / Transolver-$\sigma$'s size; $\dagger$ = count not reported, ours from the paper.
  Published rows other than Transolver-$\sigma$'s own are retrainings reported by \citet{transolversigma2026} (EddyFormer \citep{du2025eddyformer}, FactFormer,
  DRIFT-Net \citep{li2025driftnet}) and by \citet{li2023factformer} (Dil-ResNet \citep{stachenfeld2022coarse}). \emph{(b)}~MHD\_64 \citep{ohana2024well}: VRMSE, mean over the seven fields. One-step:
  all windows of the test trajectories. The Well's protocol: rollouts from the first four frames, mean over predicted
  steps 6--12 and 13--30 and over trajectories; CNextU-net \citep{liu2022convnext} also retrained by us with The Well's code and budget.
  Walrus' protocol \citep{mccabe2025walrus}: rollouts from frames 7--10, median over trajectories of the mean over predicted steps 1--20 and
  21--60. EddyFormer follows The Well's protocol. Small = at CNextU-net's size.}
  \label{tab:benchmarks}
  \centering
  \footnotesize
  \setlength{\tabcolsep}{3pt}
  \begin{tabular}[t]{@{}l r r r@{}}
\multicolumn{4}{@{}c@{}}{\textbf{(a)} Isotropic turbulence} \\[2pt]
\toprule
& & \multicolumn{2}{c@{}}{$L_2$ error (\%)} \\
\cmidrule(l){3-4}
& Params (M) & $\mathbf{u}$ & $p$ \\
\midrule
\multicolumn{4}{@{}l}{\emph{Ours}} \\
UNet & 9.88 & \textbf{6.54} & \textbf{9.88} \\
UNet, small & 4.93 & 7.67 & 11.49 \\
UNet, tiny & 1.68 & 9.43 & 14.35 \\
Transformer, $\grid{64}$ & 66.27 & 8.54 & 14.02 \\
\midrule
\multicolumn{4}{@{}l}{\emph{Published}} \\
Transolver-$\sigma$ (preprint) & 2.0--2.2$\smash{^{\dagger}}$ & 10.80 & 15.65 \\
Dil-ResNet & 6.9 & 11.43 & 17.25 \\
EddyFormer & -- & 12.70 & 19.43 \\
FactFormer & 5.1 & 13.87 & 21.74 \\
DRIFT-Net & -- & 14.56 & 22.64 \\
\bottomrule
\end{tabular}
\unskip\hfill
  \begin{tabular}[t]{@{}l r r r r r r@{}}
\multicolumn{7}{@{}c@{}}{\textbf{(b)} MHD\_64, VRMSE} \\[2pt]
\toprule
& & & \multicolumn{2}{c}{The Well} & \multicolumn{2}{c@{}}{Walrus} \\
\cmidrule(lr){4-5} \cmidrule(l){6-7}
& Params (M) & One-step & 6--12 & 13--30 & 1--20 & 21--60 \\
\midrule
\multicolumn{7}{@{}l}{\emph{Ours}} \\
UNet & 37.93 & 0.140 & 0.718 & 1.041 & 0.701 & 1.134 \\
UNet, small & 21.48 & 0.154 & 0.720 & \textbf{1.026} & 0.707 & \textbf{1.101} \\
Transformer & 66.30 & 0.161 & \textbf{0.689} & 1.031 & 0.683 & 1.156 \\
Transformer, small & 21.09 & 0.177 & 0.694 & 1.030 & 0.690 & 1.151 \\
\midrule
\multicolumn{7}{@{}l}{\emph{Published}} \\
CNextU-net & 22.3 & 0.1633 & 1.30 & 2.23 & -- & -- \\
CNextU-net, retrained & 22.3 & 0.153 & 1.18 & 1.85 & -- & -- \\
FNO & 302 & 0.3605 & 1.24 & 1.61 & -- & -- \\
EddyFormer & -- & 0.1160 & -- & -- & -- & -- \\
Walrus & 1300 & \textbf{0.0580} & -- & -- & \textbf{0.6487} & 1.2256 \\
\bottomrule
\end{tabular}

\end{table*}

\begin{figure*}[t]
  \centering
  \includegraphics[width=\textwidth]{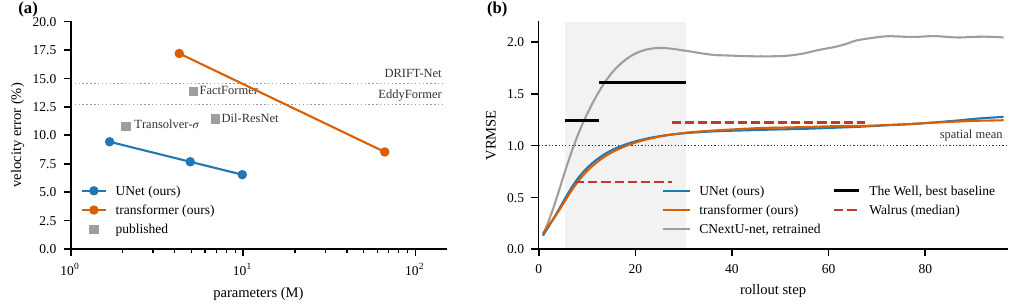}
  \caption{\emph{(a)}~Isotropic turbulence: velocity error against parameter count for our UNet (three widths) and
  transformer (two widths; the 4\,M model is tabulated in App.~\ref{app:protocols}) and for the published models (grey;
  dotted lines: no reported count; bar: the 2.0--2.2\,M range of our count for Transolver-$\sigma$). \emph{(b)}~MHD\_64:
  VRMSE per rollout step, averaged over fields, the ten test trajectories and three seeds, for our UNet (38\,M) and
  transformer (66\,M) and for the retrained CNextU-net (grey). Shaded: The Well's windows; black: the published means of
  its best baseline (FNO) over them; red, dashed: Walrus' published medians over its windows, drawn at their frames (its
  rollouts start at frame 11, our step 8).}
  \label{fig:benchmarks}
\end{figure*}
\section{Public Benchmarks}
\label{sec:benchmarks}

The models and the training of \cref{sec:method} were designed for the decaying MHD turbulence of \cref{sec:mhd128}. To measure how well they transfer and to compare them with published models, we apply the same backbones and two-stage
training, adapted to each benchmark's fields, grid and input length, to two public
benchmarks of three-dimensional turbulence that come with their own data, splits, metrics and published results: the
isotropic turbulence set of FactFormer \citep{li2023factformer} and the MHD data set of The Well \citep{ohana2024well}.
Both flows are forced rather than decaying, and the second is compressible and strongly magnetized. We tuned little.
The learning rate was chosen among three or four values and the number of input frames among 4, 6 and 10, on the validation
splits. The remaining settings are those of \cref{tab:settings}, with the stage lengths set per data set by a validation rule
(App.~\ref{app:protocols}). Every rule was fixed before the runs it governs, every choice was made on validation, and each
test split was scored once, with the final models. Rollouts start from the given frames of a test trajectory and are
never re-initialized from the reference. Each final model was trained on one GPU in at most \resBenchHoursMax{}
GPU-hours, with three seeds. \Cref{tab:benchmarks} and \cref{fig:benchmarks} hold the results. On the isotropic set
our models set the lowest errors reported to date, by a wide margin and down to 1.7\,M parameters; on MHD\_64 they reach
rollout errors a third to a half below the best of The Well's baselines and match a 1.3\,B-parameter pretrained foundation model
in its rollout windows.

\subsection{Isotropic turbulence}
\label{sec:bench:iso}

The benchmark is forced, statistically stationary, incompressible turbulence in a periodic box, stored on a $\grid{60}$
grid at $\Rel = 84$ \citep{li2023factformer}, the range of our $\Re = 6000$ data. From the first 10 frames of a test
trajectory a model predicts velocity and pressure over the next 10. The error of a field is its relative $L_2$ norm over
space and the 10 predicted frames, averaged over the 100 test trajectories. The velocity error is the mean of the errors
of its three components.

Our backbones need a grid that can be halved more often than $\grid{60}$ allows, so each is adapted. The UNet uses three
levels instead of four. The transformer keeps its $\grid{128}$ layout and runs on a $\grid{64}$ grid, with each input
zero-padded in Fourier space and each prediction truncated back, a round trip that is the identity on the data grid. Both
models also predict the pressure as an additional field, take the last four of the given frames instead of two, and drop
the time channel because the flow is stationary. The predicted velocity is projected as in \cref{eq:rollout}, but not
clipped.

The UNet with \resIsoMainParams\,M parameters reaches \resIsoMainU\,\% / \resIsoMainP\,\% (velocity / pressure), the
lowest error reported on this benchmark to date. It is \resIsoMainBelowTransolverU\,\% / \resIsoMainBelowTransolverP\,\% below
Transolver-$\sigma$ \citep[preprint;][]{transolversigma2026}, the best published model, and \resIsoMainBelowDilU\,\%
below the dilated ResNet of FactFormer's own comparison, the best
peer-reviewed one. The lead does not come from model size (\cref{fig:benchmarks}a). At FactFormer's size the UNet reaches
\resIsoSmallU{} / \resIsoSmallP, and with \resIsoTinyParams\,M parameters, at or below Transolver-$\sigma$'s count, it
reaches \resIsoTinyU{} / \resIsoTinyP, still ahead of every published model. The transformer reaches \resIsoXU{} /
\resIsoXP{} with 66\,M parameters, also below every published model. Its 4\,M version does not
(\cref{tab:bench_iso_full}), so on this short-horizon task the UNet is the better small model. Seeds move the UNet rows
by at most \resIsoUnetMaxSd{} points. The
one-step stage of the UNet rows ended at its cap with the validation error still falling (App.~\ref{app:protocols}), so
these numbers are not the limit of the approach on this set.

Under the same protocol, the U-Net retrained by \citet{transolversigma2026} scores
\pubIsoUnetU{} / \pubIsoUnetP{}. To locate what our model adds, we changed one element at a time of the UNet
at FactFormer's size, keeping its seeds, data and evaluation, and fixed the thresholds for a meaningful change before the runs
(\cref{tab:bench_iso_abl}). The training matters most. Without the rollout stage, the velocity error rises by
\resIsoAblOneStepDeltaU\,\%. Rollout training alone, from random initialization and at the same GPU time, raises it by
\resIsoAblScratchDeltaU\,\%. Switching off the input noise, giving two input frames instead of four, or removing the
self-attention at the coarsest level changes it by at most \resIsoAblNullMaxAbsDeltaU\,\% on this ten-frame horizon. Every variant stays below every
published model, at \resIsoAblMaxU{} / \resIsoAblMaxP{} or less, so no single one of these elements accounts for the lead.

\subsection{Compressible MHD turbulence}
\label{sec:bench:mhd}

MHD\_64 is the MHD data set of The Well \citep{ohana2024well}, from the simulation series of
\citet{cho2003compressible,burkhart2009density,portillo2018threepoint,burkhart2020cats}: isothermal, compressible, ideal
MHD turbulence, continuously driven, simulated at $\grid{256}$ and filtered to $\grid{64}$, in ten regimes that span sonic
Mach numbers from 0.5 to 7 and sub- and super-Alfv\'enic fields. Its 100 trajectories (97 in the released files) of 100 frames hold density,
velocity and magnetic field; the official split is 77 / 10 / 10. The flow differs from our data in forcing, compressibility and field strength. It is driven,
shock-dominated at the high Mach numbers, and carries a magnetic energy of 0.3 to 2.8 times the kinetic energy,
at $\Rel \approx 55$--$80$ (App.~\ref{app:protocols}). The metric is VRMSE, the root mean squared error divided by the
spatial standard deviation of the truth (1 = predicting the spatial mean), averaged over the seven fields. The Well
reports the one-step error over all windows of the test trajectories and rollouts from the first four frames, averaged
over predicted steps 6--12 and 13--30. Its four baselines were trained for one step, 12 hours each on one H100, and
the FNO has the lowest rollout errors among them. Because the released checkpoints do not reproduce the published
numbers (App.~\ref{app:protocols}), we also retrained the strongest one-step baseline, CNextU-net
\citep{liu2022convnext}, with The Well's current code and budget. Walrus \citep{mccabe2025walrus}, a 1.3\,B-parameter
transformer pretrained on 19 data sets including MHD\_64 and fine-tuned on it, reports the lowest errors on this set
under its own protocol: rollouts whose first predicted frame is frame 11, the median over trajectories of the mean over
predicted steps 1--20 and 21--60. We evaluate every model under both protocols. Our models carry the seven fields, four
input frames and no time channel, without projection or clipping. The loss divides the squared error of each field by
its spatial variance, the training-time form of VRMSE, because the field variances differ by orders of magnitude
between the regimes.

In rollouts our models reach \resMhdWinAMin--\resMhdWinAMax{} over steps 6--12 and \resMhdWinBMin--\resMhdWinBMax{} over
steps 13--30, at 21\,M parameters as at 38 or 66\,M, with standard deviations over seeds of at most \resMhdMaxSdWin.
These values are \resMhdBelowFnoWinAMin--\resMhdBelowFnoWinAMax\,\% and \resMhdBelowFnoWinBMin--\resMhdBelowFnoWinBMax\,\%
below the best published ones, and \resMhdBelowRetrainWinAMin--\resMhdBelowRetrainWinAMax\,\% and
\resMhdBelowRetrainWinBMin--\resMhdBelowRetrainWinBMax\,\% below the retrained CNextU-net. The per-step curves (\cref{fig:benchmarks}b)
show why. The error of our models grows more slowly. It crosses VRMSE 1 at step
\resMhdStepAboveOneMin--\resMhdStepAboveOneMax{}, where the retrained CNextU-net crosses it at step
\resCnextRetrainStepAboveOne, and settles at \resMhdStepLastMin--\resMhdStepLastMax{} by step 96 against
\resCnextRetrainStepLast. Under Walrus' protocol our models are within
\resMhdVsWalrusEarlyMin--\resMhdVsWalrusEarlyMax\,\% of Walrus over steps 1--20 and
\resMhdVsWalrusLateMin--\resMhdVsWalrusLateMax\,\% below it over steps 21--60, with
\resMhdWalrusRatioMin--\resMhdWalrusRatioMax{} times fewer parameters and trained from scratch on this data set alone.
Past about 20 steps every curve in \cref{fig:benchmarks}b is above VRMSE 1, and Walrus' own mean over steps 21--60 is
\pubMhdWalrusWalrusLate. The predicted trajectories have thus decorrelated from the reference, and, as
\citet{mccabe2025walrus} note for chaotic systems, a pointwise error then saturates and no longer separates models. The one-step error,
which the training targets only in its first stage, is at the level of CNextU-net
(\resMhdOneStepMin--\resMhdOneStepMax) and above EddyFormer \citep[\pubMhdEddyformerOneStep;][]{du2025eddyformer} and
Walrus (\pubMhdWalrusOneStep).\footnote{A further MHD\_64 result \citep{scalesplit2026} uses one of the ten regimes, an
error normalized by the field's norm rather than its variance, and two-step rollouts restarted from the reference; it is
not comparable with the numbers here and is not tabulated.} The approach, developed for decaying incompressible MHD,
thus carries over to forced hydrodynamic turbulence and to compressible, strongly magnetized turbulence without
re-design, at a few GPU-hours per model.

\section{Discussion and Limitations}
\label{sec:discussion}
Two standard backbones, trained with one procedure and one stopping rule, follow decaying three-dimensional MHD turbulence
at $\grid{128}$ for about one large-eddy turnover time with velocity errors of a few per cent at three Reynolds numbers, for
example \resMReThreeKUnetU{} for the UNet at $\Re = 3000$ and \resMReOneKXfU{} for the transformer at $\Re = 1000$
(\cref{tab:mhd128}). The predictions retain the small scales. The current sheets are in place, the enstrophy and the mean
squared current density stay within a few per cent of the truth's, and the energy spectra match the truth's over the
wavenumbers that hold nearly all of the energy, departing only in the dissipative tail near the grid scale
(\cref{sec:mhd128:results}). Which of the two is the more accurate depends on the regime:
the UNet leads at $\Re = 3000$ and, by a smaller margin, at $\Re = 6000$, where the transformer comes within
\resMReSixKXfAboveUnetMaxPct{}\,\% of it in \resMReSixKXfHours{} GPU-hours against \resMReSixKUnetHours{}, and the
transformer leads at $\Re = 1000$; the FNO trails
both at every Reynolds number. The ranking is conditional on equal training data and one stopping rule. The same backbones and two-stage training, adapted to each benchmark's fields, grid and input length, transfer to forced
hydrodynamic
turbulence and to compressible, strongly magnetized MHD (\cref{sec:benchmarks}). The models reach the lowest errors reported to date on
FactFormer's isotropic turbulence set and, on MHD\_64, rollout errors a third to a half below the best of The Well's baselines from 77
training trajectories, at a few GPU-hours per model.

\paragraph{Limitations.}
Our rollouts cover one stored window, $T = 25$ steps or about one large-eddy turnover time. By its end the decaying flow
retains \resMReOneKEkinEndPct{}--\resMReSixKEkinEndPct{}\,\% of its kinetic energy and half of its $\Rel$
(\cref{sec:mhd128:data}), and a longer window of these flows would mainly record their viscous decay, so longer horizons are
tested on the forced, statistically stationary benchmarks (\cref{sec:benchmarks}). The models have headroom at the highest
Reynolds number. The UNet's validation error at $\Re = 6000$ was still falling when the stopping rule ended its training,
and its error on held-out simulations exceeds that on the training simulations (\cref{tab:mhd128_runs}), the signature of
a fit limited by the 400 training simulations rather than by the model. More simulations, or a data-scaling study, are the
direct next step there. The evidence for the order of the two training stages is one controlled test on the UNet at
$\Re = 3000$ (App.~\ref{app:training}). The natural next steps are a forced, statistically stationary MHD set with a
dynamically significant magnetic field, on which the same models can be tested over many turnover times, and, beyond the
predictability time, ensembles or generative models in place of a single deterministic rollout.

We will release the code, the configuration of every model, the evaluation scripts, and the data generator with the seed
of every simulation upon publication, so that the three $\grid{128}$ data sets can be regenerated; the test simulations are available on
request (App.~\ref{app:repro}).

  \section*{Acknowledgements}
E.A.H. acknowledges support from NSF grants \mbox{OAC-2514142} and \mbox{OAC-2209892}.
This research used both the DeltaAI advanced computing and data resource, which is supported by the National Science Foundation
(award OAC~2320345) and the State of Illinois, and the Delta advanced computing and data resource which is supported by the
National Science Foundation (award OAC~2005572) and the State of Illinois. Delta and DeltaAI are joint efforts of the University of
Illinois Urbana-Champaign and its National Center for Supercomputing Applications.
This work used the Delta and DeltaAI systems at the National Center for Supercomputing Applications through allocations
\mbox{CIS261641} and \mbox{CIS250449} from the Advanced Cyberinfrastructure Coordination Ecosystem: Services \& Support (ACCESS)
program, which is supported by National Science Foundation grants \#2138259, \#2138286, \#2138307, \#2137603, and \#2138296.
Access to DeltaAI was provided in part by the Illinois Computes project which is supported by the University of Illinois
Urbana-Champaign.

The authors used generative AI tools to assist with writing, code and figures. The authors decided the research ideas, model
designs and conclusions on the basis of their experiments. They reviewed the AI-assisted content and take full responsibility for
this paper.

\section*{Impact Statement}
This work advances the learned emulation of three-dimensional MHD turbulence, the flow that shapes the solar wind and the
interstellar gas in which stars form; the same equations govern the edge of fusion plasmas. Simulations of this flow are expensive enough that
parameter surveys, design and control studies and inference from observations are limited by computing time. Surrogates
of the accuracy shown here, trained in tens of GPU-hours, can lower that barrier for astrophysics,
space physics and fusion research. The code, configurations and data generator, to be released upon publication, will let
others reproduce and extend the study. The models are trained on idealized simulations over short horizons and are not validated for operational
use; their predictions outside the tested regimes should be checked against a solver. We see no societal consequences
that must be specifically highlighted beyond those of scientific machine learning in general.

\bibliography{references}
\bibliographystyle{icml2026}

\newpage
\appendix
\onecolumn
\section{Data}
\label{app:data}

\paragraph{Simulations.}
Each data set is 500 simulations of \cref{eq:mhd} with $\nu = \eta = 1/\Re$ in the periodic unit box, run with Dedalus~3
\citep{burns2020dedalus} on $\grid{128}$ Fourier modes with $3/2$ dealiasing and a four-stage third-order implicit--explicit
Runge--Kutta scheme (viscous and resistive terms implicit, the rest explicit) at a fixed step of $10^{-3}$, so that the
stored window $\tau \in [0, 1]$ is 1,000 solver steps and every 40th step is kept. The solver advances $\mathbf{u}$ and the
vector potential $\mathbf{A}$; the stored channels are $\mathbf{u}$ and $\mathbf{B} = \nabla\times\mathbf{A}$, the curl taken
spectrally. Each simulation ran on \resMReThreeKSolverRanks{} MPI ranks and took a median of \resMReThreeKSolverSecPerStep{}\,s per step at $\Re = 3000$,
\resMReThreeKSolverWallHoursPerSim{} hours of wall time or \resMReThreeKSolverCpuHoursPerSim{} CPU-hours per simulation
(\resMReOneKSolverSecPerStep{} and \resMReSixKSolverSecPerStep{}\,s per step at $\Re = 1000$ and $6000$).

\paragraph{Initial conditions and checks.}
Simulation $i$ of the set at Reynolds number $\Re$ draws its initial condition from the seed $1000\,\Re + i$. Every
component of $\boldsymbol{\Phi}$ and $\mathbf{A}_0$ (\cref{sec:mhd128:data}) is a periodic Gaussian random field with
squared-exponential covariance of correlation length $0.1$ and standard deviation $0.1$ for $\boldsymbol{\Phi}$ and
$5\times10^{-3}$ for $\mathbf{A}_0$; the curls $\mathbf{u}_0 = \nabla\times\boldsymbol{\Phi}$ and
$\mathbf{B}_0 = \nabla\times\mathbf{A}_0$ are divergence-free by construction. Every stored value is finite; no two initial
conditions within a set or across sets are near-duplicates; the relative divergence of
the stored velocity is below $3\times10^{-7}$; and the kinetic energy decays in every simulation.

\paragraph{Splits and normalization.}
The 500 indices of a set are permuted once with NumPy's default generator seeded with 42; the first 400 are the training
simulations, the next 50 the validation and the last 50 the test simulations, the same rule at every Reynolds number. Per-channel
normalization statistics are computed on the training split only.

\paragraph{Flow statistics.}
With $\hat{f}(\mathbf{m})$ the Fourier coefficient at integer mode $\mathbf{m}$ and $\mathbf{k} = 2\pi\mathbf{m}$, the
kinetic energy is $E_{\mathrm{kin}} = \tfrac12\langle|\mathbf{u}|^2\rangle$, the dissipation rate
$\varepsilon = \nu\langle|\nabla\mathbf{u}|^2\rangle = \nu\sum_{\mathbf{m}}|\mathbf{k}|^2|\hat{\mathbf{u}}(\mathbf{m})|^2$,
the root-mean-square velocity component $u' = (2E_{\mathrm{kin}}/3)^{1/2}$, the Taylor microscale
$\lambda = (15\nu u'^2/\varepsilon)^{1/2}$ and $\Rel = u'\lambda/\nu$ \citep[Sec.~6.1 and 6.3]{pope2000turbulent}; the
Kolmogorov scale is $\eta = (\nu^3/\varepsilon)^{1/4}$ and $k_{\max} = 2\pi \cdot 64$ the largest resolved wavenumber. The
integral scale $L = 3\pi/(4E_{\mathrm{kin}}) \sum_n E(n)/k_n$ from the shell-summed spectrum $E(n)$
\citep[Eq.~6.260]{pope2000turbulent} gives the large-eddy turnover time $L/u'$, and the number of turnover times in the
window is its length divided by the window mean of $L/u'$. All quantities are computed per frame on the first 16 training
simulations of each set; \cref{tab:data_stats} lists their frame means. The grid resolves the Kolmogorov scale at every Reynolds
number, since the smallest $k_{\max}\eta$ over the window is \resMReSixKKmaxEtaMin{} at $\Re = 6000$, and the spectra show no
pile-up at the grid scale.

\begin{table}[!h]
  \caption{Statistics of the stored windows ($\tau \in [0, 1]$; frame means over the first 16 training simulations of each set).
  $\Rel$ at the first and last frame; kinetic energy at $\tau = 1$ relative to $\tau = 0$; large-eddy turnover times
  in the window; smallest value over the window of the largest resolved wavenumber times the Kolmogorov scale.}
  \label{tab:data_stats}
  \centering
  \footnotesize
  \setlength{\tabcolsep}{4pt}
  \begin{tabular}{@{}l r r r@{}}
    \toprule
    & $\Re = 1000$ & $\Re = 3000$ & $\Re = 6000$ \\
    \midrule
    $\Rel$, $\tau = 0 \to 1$ & \resMReOneKRelStart{} $\to$ \resMReOneKRelEnd{} & \resMReThreeKRelStart{} $\to$ \resMReThreeKRelEnd{} & \resMReSixKRelStart{} $\to$ \resMReSixKRelEnd{} \\
    $E_{\mathrm{kin}}(1) / E_{\mathrm{kin}}(0)$ (\%) & \resMReOneKEkinEndPct & \resMReThreeKEkinEndPct & \resMReSixKEkinEndPct \\
    turnover times in the window & \resMReOneKTurnovers & \resMReThreeKTurnovers & \resMReSixKTurnovers \\
    $\min_\tau\, k_{\max}\eta$ & \resMReOneKKmaxEtaMin & \resMReThreeKKmaxEtaMin & \resMReSixKKmaxEtaMin \\
    \bottomrule
  \end{tabular}
\end{table}

\section{Architectures and Training}
\label{app:training}

\paragraph{Optimization.}
All models use AdamW with weight decay $10^{-4}$ and $(\beta_1, \beta_2) = (0.9, 0.99)$, a linear warm-up over 2,604 updates
to a constant learning rate, gradient clipping to norm 1, and an exponential moving average of the weights with a half-life
of 2,604 samples, which is the model evaluated. The learning rate is $4\times10^{-4}$ for the UNet and the transformer and
$1.6\times10^{-3}$ for the FNO. For the transformer, a check over $2\times10^{-4}$ to $3.2\times10^{-3}$ in factors of two on the $\grid{64}$ data at
$\Re = 3000$ put the optimum at $4\times10^{-4}$ to $8\times10^{-4}$; for the FNO, a check over $8\times10^{-4}$ to
$6.4\times10^{-3}$ on the $\grid{64}$ data at $\Re = 1000$ put it at $1.6\times10^{-3}$. Every update sees a
global batch of four samples, split over one, two or four GH200 GPUs by data parallelism, so that the update is the same whatever
the GPU count. The one-step stage is 32 epochs of 1,250 updates; rollout blocks are 24 and 12 epochs of 550 updates on
four-step windows. The UNet's rollout stage recomputes the first step of each window in the backward pass and back-propagates
the four windows of a batch one at a time (\cref{sec:method:training}); the other two backbones fit without. The transformer
shifts each training window cyclically in the rollout stage (\cref{sec:method:backbones}). Each model is a single training run.

\paragraph{Runs.}
\Cref{tab:mhd128_runs} lists every $\grid{128}$ model: the updates of each stage up to the reported model, the rollout
blocks trained including the block that triggered the stopping rule, the validation gain of that block (\cref{eq:gain}), the
GPU-hours of the reported model and of all blocks, and the gap between the error on the validation and on 12 training
simulations. The stopping rule uses 12 validation and 12 training simulations. The GPU-hours are the sum of the training segments' wall time times their GPU count, in-training validation
included and queueing excluded; a rollout update of the UNet costs \resMReThreeKUnetRolloutUpdateCost{} times its one-step
update, since it back-propagates through the four steps of a window and recomputes the first.
\begin{table*}[t]
  \caption{Training of the $\grid{128}$ models: parameters, updates of the one-step stage and of the rollout stage up to the
  reported model, rollout blocks trained (including the stop block), index of the reported block, the validation gain of the
  stop block over it (the stopping rule's gain, \cref{eq:gain}, at $\tau = \resMTauEnd$; the rule stops below 5\%, or,
  marked $\dagger$, when the training-simulation error falls while the validation error lags),
  GPU-hours of the reported model and of all blocks, and the gap between the validation and the training-simulation error
  of the reported model (validation / training $- 1$).}
  \label{tab:mhd128_runs}
  \centering
  \footnotesize
  \setlength{\tabcolsep}{3.5pt}
  \begin{tabular}{@{}l r r r r r r r r r@{}}
\toprule
& & \multicolumn{2}{c}{updates} & \multicolumn{2}{c}{blocks} & stop gain & \multicolumn{2}{c}{GPU-h} & gap (\%) \\
\cmidrule(lr){3-4} \cmidrule(lr){5-6} \cmidrule(lr){8-9}
& Params (M) & one-step & rollout & trained & reported & (\%) & reported & all & $\mathbf{u}$ / $\mathbf{B}$ / $\boldsymbol{\omega}$ / $\mathbf{J}$ \\
\midrule
\multicolumn{10}{@{}l}{\emph{$\Re = 1000$ ($\Rel = 9$--$17$)}} \\
UNet & 37.7 & 40,000 & 0 & 1 & 0 & 12.4$^{\dagger}$ & 20.0 & 49.9 & 190 / 76 / 170 / 107 \\
Transformer & 66.3 & 40,000 & 26,400 & 4 & 3 & 2.7 & 30.9 & 36.0 & 50 / 46 / 25 / 30 \\
FNO & 37.8 & 40,000 & 13,200 & 2 & 1 & 3.6 & 17.5 & 20.7 & 61 / 70 / 31 / 38 \\
\midrule
\multicolumn{10}{@{}l}{\emph{$\Re = 3000$ ($\Rel = 25$--$50$)}} \\
UNet & 37.7 & 40,000 & 19,800 & 3 & 2 & -1.1 & 65.0 & 80.5 & 11 / 10 / 8 / 9 \\
Transformer & 66.3 & 40,000 & 13,200 & 2 & 1 & -11.1 & 21.3 & 26.5 & 26 / 26 / 18 / 21 \\
FNO & 37.8 & 40,000 & 13,200 & 2 & 1 & 1.0 & 17.7 & 21.5 & 28 / 35 / 17 / 20 \\
\midrule
\multicolumn{10}{@{}l}{\emph{$\Re = 6000$ ($\Rel = 44$--$101$)}} \\
UNet & 37.7 & 40,000 & 19,800 & 3 & 2 & 2.7 & 64.8 & 79.7 & 48 / 37 / 33 / 32 \\
Transformer & 66.3 & 40,000 & 33,000 & 5 & 4 & 4.9 & 35.8 & 40.7 & 16 / 15 / 11 / 11 \\
FNO & 37.8 & 40,000 & 13,200 & 2 & 1 & -0.2 & 17.3 & 20.6 & 25 / 20 / 12 / 9 \\
\bottomrule
\end{tabular}

\end{table*}

\paragraph{Training cost and the two-stage procedure.}
\label{sec:mhd128:cost}
At $\Re = 3000$ the two leading backbones reach the same error at about the same training time, but at different stages of
the procedure. After its one-step stage alone (\resMReThreeKUnetStageOneHours{}
GPU-hours) the UNet reaches \resMReThreeKUnetStageOneU{} in velocity and \resMReThreeKUnetStageOneJ{} in current density.
The transformer's final model, its one-step stage (\resMReThreeKXfStageOneHours{} GPU-hours, \resMReThreeKXfStageOneU{} and
\resMReThreeKXfStageOneJ{}) followed by one rollout block, reaches \resMReThreeKXfU{} and \resMReThreeKXfJ{} at
\resMReThreeKXfHours{} GPU-hours, within \resMReThreeKUnetStageOneVsXfMaxPct{}\,\% of the UNet's stage-one model in every
quantity. The UNet's margin in \cref{tab:mhd128} comes from its own rollout training, which lowered its errors by \resMReThreeKUnetRolloutGainUPct{}\,\% in velocity and
\resMReThreeKUnetRolloutGainJPct{}\,\% in current density over two blocks, whereas the transformer's second block raised its
errors by \resMReThreeKXfStopBlockRiseMinPct{}--\resMReThreeKXfStopBlockRiseMaxPct{}\,\% and the rule ended its training
after one (\cref{tab:mhd128_runs}). At $\Re = 6000$ the transformer kept improving for four blocks and came within
\resMReSixKXfAboveUnetMinPct{}--\resMReSixKXfAboveUnetMaxPct{}\,\% of the UNet's errors at \resMReSixKXfHours{} GPU-hours
against the UNet's \resMReSixKUnetHours{}, because a rollout update costs
\resMReSixKUnetRolloutUpdateCost{} times a one-step update for the UNet and only \resMReSixKXfRolloutUpdateCost{} times for the
transformer (\cref{tab:mhd128_runs}). The order of the two stages matters. In a development test at equal GPU time
(\resMRecipeTwoStageHours{} hours; $\Re = 3000$, UNet), rollout training from random initialization ended \resMRecipeScratchOverU{}, \resMRecipeScratchOverB{},
\resMRecipeScratchOverW{} and \resMRecipeScratchOverJ{} times worse in velocity, magnetic field, vorticity and current
density than one-step training followed by rollout training, in every simulation compared
(\cref{tab:mhd128_recipe}).

\paragraph{Development tests.}
The configuration of each backbone was fixed before the runs of \cref{tab:mhd128} by a small search at $\grid{128}$ and
$\Re = 3000$, judged on 12 validation simulations after the one-step stage; a variant replaced the base configuration only if it was at least 10\,\% better at
equal GPU time and at most $1.3$ times as expensive per update. For the UNet (width 56, five levels, three residual blocks
per level, window attention at the $\grid{32}$ and $\grid{16}$ levels) and for the FNO (widths and mode counts around the
base) no variant met the bar; for the transformer the search fixed the width at 128.

\paragraph{Choices tested.}
Three settings of \cref{tab:settings} were fixed by one development run each, all at $\Re = 3000$. The length of the
one-step stage: on the UNet at $\grid{128}$, 80,000 one-step updates followed by 13,200 rollout updates lowered the
current-density error by 8\,\% (in 11 of 12 validation simulations) against 40,000 followed by the same rollout updates,
left velocity and vorticity unchanged, and cost about 1.4 times as much (75 against 52 GPU-hours), so $U_1 = 40{,}000$ was
kept for every backbone. The input-noise range: $[a_{\min}, a_{\max}] = [0.01, 0.3]$ was set so that $a\,\sigma_{\mathrm{d}}$
spans the per-voxel errors of rollout inputs measured in development, from about $0.002$ after the first step to about
$0.04$ at the end of the window in normalized units. The shifted training windows: on the $\grid{64}$ data, shifting lowered
the transformer's errors at the end of the window by 15\,\% in velocity (in 12 of 12 simulations), 8\,\% in vorticity and
4\,\% in current density, and the UNet's by 1\,\%, within the run-to-run spread, so the transformer alone uses it.
\begin{table}[t]
  \caption{Development test of the two-stage procedure at equal GPU time (Re $= 3000$, UNet, 12 validation simulations,
  $\tau = \resMTauEnd$): two-stage training against rollout
  training from random initialization.}
  \label{tab:mhd128_recipe}
  \centering
  \footnotesize
  \setlength{\tabcolsep}{4pt}
  \begin{tabular}{@{}l r r r r r r@{}}
\toprule
& GPU-h & updates & $\mathbf{u}$ & $\mathbf{B}$ & $\boldsymbol{\omega}$ & $\mathbf{J}$ \\
\midrule
two-stage & 41 & 40,000 + 8,800 & 0.030 & 0.077 & 0.064 & 0.173 \\
rollout training from scratch & 41 & 17,600 & 0.039 & 0.116 & 0.096 & 0.378 \\
\midrule
ratio from scratch / two-stage & & & 1.32 & 1.50 & 1.50 & 2.18 \\
from scratch worse in (of 12 simulations) & & & 12 & 12 & 12 & 12 \\
\bottomrule
\end{tabular}

\end{table}

\section{Benchmark Protocols}
\label{app:protocols}

\Cref{tab:bench_protocols} states the protocol of each benchmark of \cref{sec:benchmarks}; \cref{tab:bench_iso_full,tab:bench_mhd_full}
hold every row with its standard deviation over seeds. Every rule was written into a dated protocol before the runs it
governs, selections were made on validation splits only, and each test split was scored once, with the final models.

\begin{table}[t]
  \caption{Benchmark protocols. Ours = the choices of our rows; the rest follows the benchmark. Cost = GPU-hours of the
  final model's stages on one GH200, in-training validation included, over all rows of
  \cref{tab:bench_iso_full,tab:bench_mhd_full}.}
  \label{tab:bench_protocols}
  \centering
  \footnotesize
  \begin{tabular}{@{}l p{0.37\textwidth} p{0.40\textwidth}@{}}
    \toprule
    & Isotropic turbulence \citep{li2023factformer} & MHD\_64 \citep{ohana2024well} \\
    \midrule
    Flow & forced incompressible turbulence, $\Rel = 84$ & driven compressible ideal MHD, ten regimes (sonic Mach 0.5--7, Alfv\'enic Mach 0.7 and 2), $\Rel \approx 55$--80 (our estimate) \\
    Grid, fields & $\grid{60}$; $\mathbf{u}$, $p$ & $\grid{64}$; $\rho$, $\mathbf{u}$, $\mathbf{B}$ \\
    Trajectories & 1,100 of 20 frames & 100 of 100 frames (97 in the released files) \\
    Split & 1,000 / 100, fixed by FactFormer's code; no validation split, ours = the last 50 training trajectories & 77 / 10 / 10, official \\
    Input frames & 10 given; ours use the last 4 & 4 (Walrus: 10 in its released MHD\_64 configuration, 3 in its paper) \\
    Metric & relative $L_2$ error (\%) per field over space and the 10 predicted frames & VRMSE per field, mean over the seven fields \\
    Rollout start & frame 10 (frames numbered from 0), never re-initialized & frame 4; Walrus' protocol: frame 11 \\
    Windows & the 10 predicted frames & predicted steps 6--12 and 13--30; Walrus' protocol: 1--20 and 21--60 \\
    Aggregation & mean over 100 test trajectories and over the velocity components & mean over 10 test trajectories; Walrus' protocol: median over the ten, the lower of the two middle values as in its released code \\
    Post-processing (ours) & divergence-free projection of $\mathbf{u}$ & none \\
    UNet levels (ours) & 3 & 4 \\
    Final models (ours) & trained on all 1,000 training trajectories with the stage lengths of a validation run; last checkpoint & stage lengths by the validation rule; best-scoring stage \\
    Stage-1 epochs, blocks & \resIsoMainStageOneEpochs{} (UNet, the cap), \resIsoXStageOneEpochs--\resIsoXSmallStageOneEpochs{} (transformer); 2 blocks & \resMhdStageOneEpochsAllMin--\resMhdStageOneEpochsAllMax{} (plateau); \resMhdBlocksMin--\resMhdBlocksMax{} blocks \\
    Cost per final model & \resIsoHoursMin--\resIsoHoursMax{} GPU-h & \resMhdHoursAllMin--\resMhdHoursAllMax{} GPU-h \\
    Published rows & Transolver-$\sigma$ (preprint) and its retrainings of EddyFormer, FactFormer and DRIFT-Net; FactFormer's own result and its Dil-ResNet & The Well's four one-step baselines (12 h on one H100 each); EddyFormer (one-step; 10\,k steps instead of the 12-h budget); Walrus, 1.3\,B parameters, pretrained on 19 data sets including MHD\_64, fine-tuned on 500\,K of its samples \\
    \bottomrule
  \end{tabular}
\end{table}

\paragraph{Training rules.}
The models are those of \cref{sec:method} with four input frames, no time channel and a residual scale per component;
the backbones keep their $\grid{128}$ configuration except the UNet's number of levels. Training follows
\cref{alg:training} with the settings of \cref{tab:settings}, except that the stage lengths are set per data set on
validation: the one-step stage runs in epochs of 1,250 updates, at most 32, and ends when the validation rollout score
has not improved by 1\,\% for four validated epochs; the rollout stage starts from the best one-step checkpoint and runs
in blocks of 2,200 updates on four-step windows until a block gains less than 5\,\%. Seeds 1--3 set the initialization
and the data order. Before the final runs, the learning rate ($1$, $2$, $4$ and $8 \times 10^{-4}$ for the UNet on the isotropic set, $2$, $4$ and
$8 \times 10^{-4}$ otherwise), the input length (4, 6 or 10 frames) and augmentation by the data's symmetries were tested
with one-step runs of 10,000 updates against a margin of 10\,\% on the validation score; only the input length on the
isotropic set passed, where four and six frames both cleared the margin and four, the better, was adopted. On that set the UNet rows reached
the cap of 32 epochs with the validation error still falling, by \resIsoMainStageOneLastGainPct\,\% over the last epoch.
Parameter counts are those of the instantiated networks.

\paragraph{Comparability.}
Our implementation of FactFormer's metric reproduces its released checkpoint through its own code to $10^{-8}$
(16.705 / 25.455, the values of its paper). Transolver-$\sigma$ does not report its parameter count on this set; counting
the architecture of its Appendix~B.3 and Table~7 with the conventions that reproduce every count the paper states gives
2.02--2.15\,M parameters with complex weights counted as two, as in its own counts (8.1--8.6\,MB), or 1.43--1.56\,M with
one per complex weight; our 1.68\,M model stores 6.7\,MB. The transformer was also run in a layout fitted to the
$\grid{60}$ grid (patch 3, three levels, window 5), which is clearly worse; both layouts are in \cref{tab:bench_iso_full}
and neither was chosen on test. On MHD\_64, The Well's paper states its rollout windows as 6--12 and 13--30 without defining the step
count. We count the first predicted frame as step 1. \Cref{tab:bench_mhd_full} also gives the windows as predicted
steps 7--12 and 14--30, the alternative reading of the bounds; every comparison of \cref{sec:benchmarks} holds under both. The released
checkpoints of The Well's baselines do not reproduce their published one-step errors with The Well's own code (its
issue 49), so a claim against them must beat the published numbers. The retrained CNextU-net of \cref{tab:benchmarks}
is The Well's current code and configuration (learning rate $5 \times 10^{-3}$, 12 h on one GPU, best-validation
checkpoint), scored with The Well's own evaluation code. Two further rows, one per backbone, were selected on validation under Walrus'
protocol among input lengths of 4, 6 and 10 frames and two schedules (four frames won); they are reported in
\cref{tab:bench_mhd_full} and lie within 0.01 of the rows of \cref{tab:benchmarks}. Rows of \citet{scalesplit2026} are not
tabulated (\cref{sec:benchmarks}).

\begin{table}[t]
  \caption{Isotropic turbulence, all rows: relative $L_2$ error (\%) on the test split, velocity / pressure, with and
  without the divergence-free projection of the predicted velocity (ours: mean $\pm$ standard deviation over three seeds;
  Transolver-$\sigma$: as reported). Small / tiny as in \cref{tab:benchmarks}; $\grid{64}$ = the native transformer layout
  on the resampled grid, $\grid{60}$ = the layout fitted to the $\grid{60}$ grid. Superscripts: p = preprint;
  r = retrained and reported by another paper; $\dagger$ = our count.}
  \label{tab:bench_iso_full}
  \centering
  \footnotesize
  \begin{tabular}{@{}l r r r r r@{}}
\toprule
& & \multicolumn{2}{c}{with projection} & \multicolumn{2}{c@{}}{without projection} \\
\cmidrule(lr){3-4} \cmidrule(l){5-6}
& params (M) & $\mathbf{u}$ & $p$ & $\mathbf{u}$ & $p$ \\
\midrule
\multicolumn{6}{@{}l}{\emph{Ours} (test, mean $\pm$ standard deviation over 3 seeds)} \\
UNet & 9.88 & 6.54 $\pm$ 0.02 & 9.88 $\pm$ 0.03 & 6.71 $\pm$ 0.02 & 9.98 $\pm$ 0.03 \\
UNet, small & 4.93 & 7.67 $\pm$ 0.03 & 11.49 $\pm$ 0.05 & 7.85 $\pm$ 0.03 & 11.61 $\pm$ 0.05 \\
UNet, tiny & 1.68 & 9.43 $\pm$ 0.01 & 14.35 $\pm$ 0.06 & 9.59 $\pm$ 0.01 & 14.43 $\pm$ 0.06 \\
Transformer, $\grid{64}$ & 66.27 & 8.54 $\pm$ 0.13 & 14.02 $\pm$ 0.34 & 8.94 $\pm$ 0.27 & 14.20 $\pm$ 0.57 \\
Transformer, $\grid{64}$, small & 4.25 & 17.19 $\pm$ 0.61 & 27.13 $\pm$ 1.21 & 17.79 $\pm$ 0.68 & 27.19 $\pm$ 1.17 \\
Transformer, $\grid{60}$ & 19.67 & 14.40 $\pm$ 0.15 & 22.05 $\pm$ 0.29 & 14.68 $\pm$ 0.14 & 22.08 $\pm$ 0.34 \\
Transformer, $\grid{60}$, small & 5.05 & 18.76 $\pm$ 0.17 & 28.65 $\pm$ 0.31 & 19.13 $\pm$ 0.20 & 28.53 $\pm$ 0.35 \\
\midrule
\multicolumn{6}{@{}l}{\emph{Published}} \\
Transolver-$\sigma$$^{\mathrm{p}}$ & 2.0--2.2$\smash{^{\dagger}}$ & 10.80 $\pm$ 0.16 & 15.65 $\pm$ 0.23 & -- & -- \\
Dil-ResNet$^{\mathrm{r}}$ & 6.9 & 11.43 & 17.25 & -- & -- \\
EddyFormer$^{\mathrm{r}}$ & -- & 12.70 & 19.43 & -- & -- \\
FactFormer$^{\mathrm{r}}$ & 5.1 & 13.87 & 21.74 & -- & -- \\
DRIFT-Net$^{\mathrm{r}}$ & -- & 14.56 & 22.64 & -- & -- \\
FactFormer & 5.1 & 16.70 & 25.45 & -- & -- \\
F-FNO$^{\mathrm{r}}$ & 3.0 & 22.98 & 31.51 & -- & -- \\
FNO$^{\mathrm{r}}$ & 509.8 & 33.82 & 46.34 & -- & -- \\
U-Net (2015)$^{\mathrm{r}}$ & -- & 35.76 & 48.49 & -- & -- \\
\bottomrule
\end{tabular}

\end{table}

\begin{table}[t]
  \caption{Isotropic turbulence, ablation of the UNet at FactFormer's size (\cref{sec:bench:iso}): relative $L_2$ error
  (\%) on the test split with the divergence-free projection, mean $\pm$ standard deviation over three seeds;
  $\Delta\mathbf{u}$ = change of the velocity error relative to the reference, the row \emph{UNet, small} of
  \cref{tab:benchmarks}. Each variant changes one element of the reference and keeps its seeds, data and evaluation; the
  learning rate is not re-tuned. No rollout stage: the reference's own model at the end of its one-step stage. Rollout
  stage only: \resIsoAblScratchUpdates{} updates on four-step windows from random initialization, the reference's
  training time. No self-attention: the attention block at the coarsest level removed. GPU-h: training of the final model
  on one GH200. Thresholds fixed before the runs: at least 10\,\% higher = the element matters; within 5\,\% = it does not.}
  \label{tab:bench_iso_abl}
  \centering
  \footnotesize
  \begin{tabular}{@{}l r r r r r@{}}
\toprule
& params (M) & $\mathbf{u}$ & $p$ & $\Delta\mathbf{u}$ (\%) & GPU-h \\
\midrule
Reference & 4.93 & 7.67 $\pm$ 0.03 & 11.49 $\pm$ 0.05 & -- & 2.5 \\
\midrule
No rollout stage & 4.93 & 8.69 $\pm$ 0.04 & 13.43 $\pm$ 0.10 & $+13.3$ & 1.7 \\
Rollout stage only, equal GPU time & 4.93 & 8.97 $\pm$ 0.02 & 13.53 $\pm$ 0.05 & $+16.9$ & 2.4 \\
No input noise & 4.93 & 7.53 $\pm$ 0.06 & 11.32 $\pm$ 0.09 & $-1.9$ & 2.4 \\
Two input frames & 4.85 & 7.85 $\pm$ 0.06 & 11.78 $\pm$ 0.15 & $+2.3$ & 2.3 \\
No self-attention & 4.81 & 7.63 $\pm$ 0.01 & 11.43 $\pm$ 0.03 & $-0.6$ & 2.3 \\
\bottomrule
\end{tabular}

\end{table}

\begin{table}[t]
  \caption{MHD\_64, all rows: test VRMSE under The Well's protocol with its windows read as predicted steps 6--12 and
  13--30 (a) or 7--12 and 14--30 (b), and under Walrus' protocol (ours: mean $\pm$ standard deviation over three seeds). Small as in \cref{tab:benchmarks}; w = selected on validation
  under Walrus' protocol; r = retrained and reported by another paper.}
  \label{tab:bench_mhd_full}
  \centering
  \scriptsize
  \setlength{\tabcolsep}{3pt}
  \begin{tabular}{@{}l r r r r r r r r@{}}
\toprule
& & & \multicolumn{2}{c}{windows (a)} & \multicolumn{2}{c}{windows (b)} & \multicolumn{2}{c@{}}{Walrus (median)} \\
\cmidrule(lr){4-5} \cmidrule(lr){6-7} \cmidrule(l){8-9}
& params (M) & one-step & 6--12 & 13--30 & 6--12 & 13--30 & 1--20 & 21--60 \\
\midrule
\multicolumn{9}{@{}l}{\emph{Ours} (test, mean $\pm$ standard deviation over 3 seeds)} \\
UNet & 37.93 & 0.140 $\pm$ 0.018 & 0.718 $\pm$ 0.007 & 1.041 $\pm$ 0.010 & 0.749 $\pm$ 0.006 & 1.050 $\pm$ 0.010 & 0.701 $\pm$ 0.011 & 1.134 $\pm$ 0.017 \\
UNet, small & 21.48 & 0.154 $\pm$ 0.003 & 0.720 $\pm$ 0.003 & 1.026 $\pm$ 0.003 & 0.750 $\pm$ 0.003 & 1.034 $\pm$ 0.004 & 0.707 $\pm$ 0.004 & 1.101 $\pm$ 0.009 \\
Transformer & 66.30 & 0.161 $\pm$ 0.021 & 0.689 $\pm$ 0.008 & 1.031 $\pm$ 0.006 & 0.720 $\pm$ 0.007 & 1.040 $\pm$ 0.006 & 0.683 $\pm$ 0.008 & 1.156 $\pm$ 0.011 \\
Transformer, small & 21.09 & 0.177 $\pm$ 0.005 & 0.694 $\pm$ 0.011 & 1.030 $\pm$ 0.003 & 0.723 $\pm$ 0.011 & 1.039 $\pm$ 0.004 & 0.690 $\pm$ 0.003 & 1.151 $\pm$ 0.027 \\
UNet$^{\mathrm{w}}$ & 37.93 & 0.127 $\pm$ 0.001 & 0.712 $\pm$ 0.003 & 1.049 $\pm$ 0.004 & 0.744 $\pm$ 0.003 & 1.057 $\pm$ 0.004 & 0.693 $\pm$ 0.003 & 1.137 $\pm$ 0.010 \\
Transformer$^{\mathrm{w}}$ & 66.30 & 0.148 $\pm$ 0.004 & 0.683 $\pm$ 0.005 & 1.033 $\pm$ 0.012 & 0.714 $\pm$ 0.005 & 1.043 $\pm$ 0.013 & 0.675 $\pm$ 0.006 & 1.157 $\pm$ 0.032 \\
\midrule
\multicolumn{9}{@{}l}{\emph{Published}, and CNextU-net retrained by us with The Well's code (one run)} \\
CNextU-net & 22.3 & 0.1633 & 1.30 & 2.23 & -- & -- & -- & -- \\
CNextU-net, retrained & 22.3 & 0.153 & 1.182 & 1.853 & 1.241 & 1.870 & -- & -- \\
U-net & 50.8 & 0.1798 & 1.65 & 4.66 & -- & -- & -- & -- \\
FNO & 302 & 0.3605 & 1.24 & 1.61 & -- & -- & -- & -- \\
TFNO & 302 & 0.3561 & 1.25 & 1.81 & -- & -- & -- & -- \\
EddyFormer & -- & 0.1160 & -- & -- & -- & -- & -- & -- \\
Walrus & 1300 & 0.0580 & -- & -- & -- & -- & 0.6487 & 1.2256 \\
DPOT-H$^{\mathrm{r}}$ & 1200 & 0.4734 & -- & -- & -- & -- & 1.0250 & 1.3055 \\
\bottomrule
\end{tabular}

\end{table}

\clearpage  %
\section{Additional Results}
\label{app:results}
\begin{figure*}[h]
  \centering
  \includegraphics[width=\textwidth]{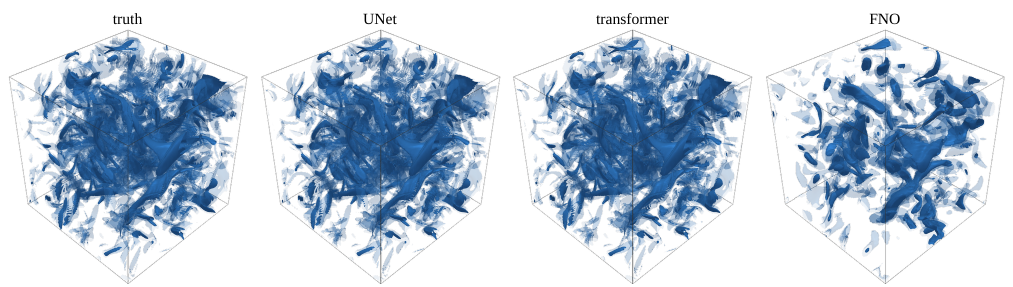}
  \caption{Magnetic field at $\tau = \resMTauEnd$, $\Re = 6000$, same simulation as \cref{fig:mhd128_slices}: isosurfaces of
  $|\mathbf{B}|$ at $0.45$ (translucent) and $0.8$ (opaque) of the truth's 99.5th percentile, the same absolute levels in
  every panel; the truth and the three models' rollouts after 25 steps.}
  \label{fig:mhd128_volumes}
\end{figure*}

\begin{figure*}[h]
  \centering
  \includegraphics{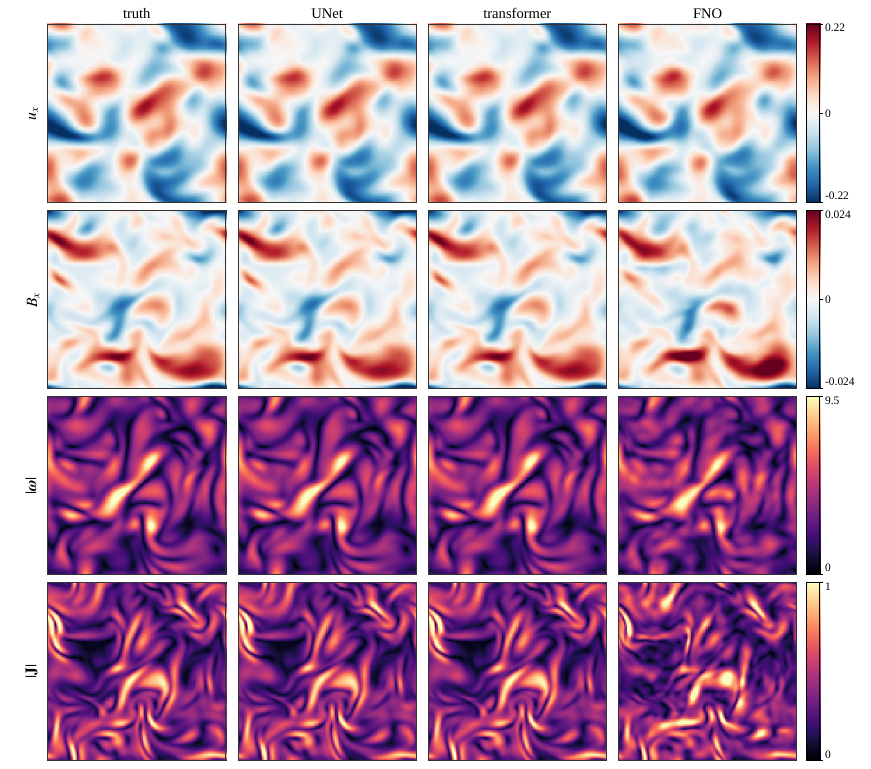}
  \caption{Mid-plane slices at $\tau = \resMTauEnd$, $\Re = 1000$, as \cref{fig:mhd128_slices}.}
  \label{fig:mhd128_slices_re1000}
\end{figure*}

\begin{figure*}[h]
  \centering
  \includegraphics{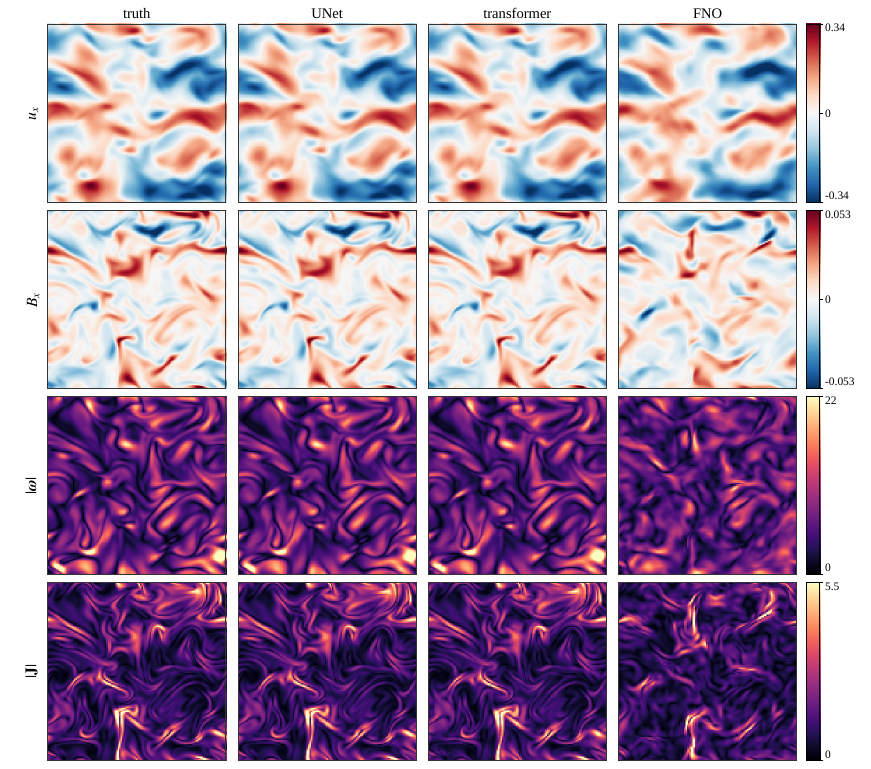}
  \caption{Mid-plane slices at $\tau = \resMTauEnd$, $\Re = 3000$, as \cref{fig:mhd128_slices}.}
  \label{fig:mhd128_slices_re3000}
\end{figure*}

This appendix collects the rollout errors at three times with their standard deviations (\cref{tab:mhd128_full}), the
sensitivity to input noise (\cref{tab:mhd128_noise}), isosurfaces of the magnetic field (\cref{fig:mhd128_volumes}), the
mid-plane slices at $\Re = 1000$ and $3000$ (\cref{fig:mhd128_slices_re1000,fig:mhd128_slices_re3000}).
\begin{table*}[h]
  \caption{Rollout errors at $\tau = \resMTauOne$, $\resMTauMid$ and $\resMTauEnd$ (steps 1, 12 and 25) on the $\grid{128}$ data
  sets (\resMSplit{} split; mean $\pm$ standard
  deviation over \resMSims{} simulations) and the ratios of predicted to true enstrophy $\Omega = \langle|\boldsymbol{\omega}|^2\rangle$
  and mean squared current density.}
  \label{tab:mhd128_full}
  \centering
  \footnotesize
  \setlength{\tabcolsep}{3.5pt}
  \begin{tabular}{@{}l r r r r r r r@{}}
\toprule
& $\tau$ & $\mathbf{u}$ & $\mathbf{B}$ & $\boldsymbol{\omega}$ & $\mathbf{J}$ & $\Omega$ ratio & $\langle|\mathbf{J}|^2\rangle$ ratio \\
\midrule
\multicolumn{8}{@{}l}{\emph{$\Re = 1000$ ($\Rel = 9$--$17$)}} \\
UNet & 0.04 & 0.004 {\scriptsize$\pm$0.001} & 0.007 {\scriptsize$\pm$0.001} & 0.010 {\scriptsize$\pm$0.001} & 0.017 {\scriptsize$\pm$0.001} & 1.000 & 1.002 \\
 & 0.48 & 0.027 {\scriptsize$\pm$0.012} & 0.034 {\scriptsize$\pm$0.011} & 0.041 {\scriptsize$\pm$0.017} & 0.055 {\scriptsize$\pm$0.018} & 0.997 & 1.000 \\
 & 1 & 0.048 {\scriptsize$\pm$0.022} & 0.068 {\scriptsize$\pm$0.021} & 0.069 {\scriptsize$\pm$0.030} & 0.094 {\scriptsize$\pm$0.034} & 0.997 & 1.001 \\
Transformer & 0.04 & 0.004 {\scriptsize$\pm$0.000} & 0.007 {\scriptsize$\pm$0.000} & 0.039 {\scriptsize$\pm$0.001} & 0.056 {\scriptsize$\pm$0.001} & 1.001 & 1.003 \\
 & 0.48 & 0.011 {\scriptsize$\pm$0.002} & 0.020 {\scriptsize$\pm$0.004} & 0.026 {\scriptsize$\pm$0.003} & 0.040 {\scriptsize$\pm$0.005} & 1.000 & 0.999 \\
 & 1 & 0.018 {\scriptsize$\pm$0.004} & 0.031 {\scriptsize$\pm$0.007} & 0.034 {\scriptsize$\pm$0.005} & 0.051 {\scriptsize$\pm$0.009} & 1.001 & 1.003 \\
FNO & 0.04 & 0.018 {\scriptsize$\pm$0.001} & 0.039 {\scriptsize$\pm$0.001} & 0.043 {\scriptsize$\pm$0.001} & 0.091 {\scriptsize$\pm$0.003} & 0.997 & 0.993 \\
 & 0.48 & 0.095 {\scriptsize$\pm$0.005} & 0.226 {\scriptsize$\pm$0.010} & 0.247 {\scriptsize$\pm$0.010} & 0.445 {\scriptsize$\pm$0.015} & 0.992 & 0.948 \\
 & 1 & 0.127 {\scriptsize$\pm$0.008} & 0.285 {\scriptsize$\pm$0.016} & 0.272 {\scriptsize$\pm$0.013} & 0.491 {\scriptsize$\pm$0.020} & 1.014 & 1.002 \\
\midrule
\multicolumn{8}{@{}l}{\emph{$\Re = 3000$ ($\Rel = 25$--$50$)}} \\
UNet & 0.04 & 0.002 {\scriptsize$\pm$0.000} & 0.006 {\scriptsize$\pm$0.001} & 0.009 {\scriptsize$\pm$0.001} & 0.026 {\scriptsize$\pm$0.003} & 1.000 & 0.998 \\
 & 0.48 & 0.012 {\scriptsize$\pm$0.004} & 0.028 {\scriptsize$\pm$0.006} & 0.031 {\scriptsize$\pm$0.008} & 0.061 {\scriptsize$\pm$0.014} & 0.998 & 0.988 \\
 & 1 & 0.029 {\scriptsize$\pm$0.012} & 0.060 {\scriptsize$\pm$0.017} & 0.062 {\scriptsize$\pm$0.022} & 0.107 {\scriptsize$\pm$0.029} & 0.997 & 0.990 \\
Transformer & 0.04 & 0.005 {\scriptsize$\pm$0.000} & 0.010 {\scriptsize$\pm$0.001} & 0.028 {\scriptsize$\pm$0.001} & 0.100 {\scriptsize$\pm$0.004} & 1.001 & 1.009 \\
 & 0.48 & 0.022 {\scriptsize$\pm$0.005} & 0.044 {\scriptsize$\pm$0.009} & 0.056 {\scriptsize$\pm$0.008} & 0.103 {\scriptsize$\pm$0.015} & 0.999 & 0.996 \\
 & 1 & 0.049 {\scriptsize$\pm$0.010} & 0.101 {\scriptsize$\pm$0.018} & 0.107 {\scriptsize$\pm$0.017} & 0.182 {\scriptsize$\pm$0.027} & 1.000 & 0.990 \\
FNO & 0.04 & 0.020 {\scriptsize$\pm$0.001} & 0.049 {\scriptsize$\pm$0.002} & 0.045 {\scriptsize$\pm$0.002} & 0.115 {\scriptsize$\pm$0.005} & 0.997 & 0.997 \\
 & 0.48 & 0.183 {\scriptsize$\pm$0.009} & 0.484 {\scriptsize$\pm$0.019} & 0.454 {\scriptsize$\pm$0.013} & 0.746 {\scriptsize$\pm$0.019} & 0.931 & 0.730 \\
 & 1 & 0.380 {\scriptsize$\pm$0.019} & 0.868 {\scriptsize$\pm$0.028} & 0.724 {\scriptsize$\pm$0.018} & 1.090 {\scriptsize$\pm$0.023} & 0.913 & 0.809 \\
\midrule
\multicolumn{8}{@{}l}{\emph{$\Re = 6000$ ($\Rel = 44$--$101$)}} \\
UNet & 0.04 & 0.002 {\scriptsize$\pm$0.000} & 0.005 {\scriptsize$\pm$0.001} & 0.009 {\scriptsize$\pm$0.001} & 0.023 {\scriptsize$\pm$0.002} & 1.000 & 1.001 \\
 & 0.48 & 0.018 {\scriptsize$\pm$0.004} & 0.046 {\scriptsize$\pm$0.007} & 0.052 {\scriptsize$\pm$0.007} & 0.112 {\scriptsize$\pm$0.015} & 0.999 & 0.978 \\
 & 1 & 0.053 {\scriptsize$\pm$0.008} & 0.127 {\scriptsize$\pm$0.018} & 0.133 {\scriptsize$\pm$0.017} & 0.245 {\scriptsize$\pm$0.033} & 0.995 & 0.974 \\
Transformer & 0.04 & 0.004 {\scriptsize$\pm$0.000} & 0.009 {\scriptsize$\pm$0.000} & 0.031 {\scriptsize$\pm$0.001} & 0.108 {\scriptsize$\pm$0.004} & 1.002 & 1.011 \\
 & 0.48 & 0.018 {\scriptsize$\pm$0.003} & 0.047 {\scriptsize$\pm$0.007} & 0.063 {\scriptsize$\pm$0.006} & 0.131 {\scriptsize$\pm$0.014} & 0.999 & 0.988 \\
 & 1 & 0.055 {\scriptsize$\pm$0.008} & 0.134 {\scriptsize$\pm$0.017} & 0.152 {\scriptsize$\pm$0.016} & 0.266 {\scriptsize$\pm$0.029} & 0.992 & 0.968 \\
FNO & 0.04 & 0.021 {\scriptsize$\pm$0.001} & 0.047 {\scriptsize$\pm$0.002} & 0.046 {\scriptsize$\pm$0.001} & 0.110 {\scriptsize$\pm$0.003} & 0.994 & 0.998 \\
 & 0.48 & 0.221 {\scriptsize$\pm$0.007} & 0.581 {\scriptsize$\pm$0.015} & 0.541 {\scriptsize$\pm$0.011} & 0.862 {\scriptsize$\pm$0.012} & 0.863 & 0.538 \\
 & 1 & 0.520 {\scriptsize$\pm$0.016} & 1.045 {\scriptsize$\pm$0.021} & 0.884 {\scriptsize$\pm$0.012} & 1.122 {\scriptsize$\pm$0.021} & 0.689 & 0.463 \\
\bottomrule
\end{tabular}

\end{table*}

\begin{table}[h]
  \caption{Sensitivity to input noise: rollouts of the final models with Gaussian noise of level $0.01$ added to the
  received frames at every step, against the noise-free rollouts, on the first four validation simulations ($\tau = \resMTauEnd$).}
  \label{tab:mhd128_noise}
  \centering
  \footnotesize
  \setlength{\tabcolsep}{4pt}
  \begin{tabular}{@{}l r r r r r r r r r@{}}
\toprule
& \multicolumn{4}{c}{noise-free} & \multicolumn{4}{c}{input noise 0.01} & \\
\cmidrule(lr){2-5} \cmidrule(lr){6-9}
& $\mathbf{u}$ & $\mathbf{B}$ & $\boldsymbol{\omega}$ & $\mathbf{J}$ & $\mathbf{u}$ & $\mathbf{B}$ & $\boldsymbol{\omega}$ & $\mathbf{J}$ & $\mathbf{J}$ ratio \\
\midrule
\multicolumn{10}{@{}l}{\emph{$\Re = 1000$ ($\Rel = 9$--$17$)}} \\
UNet & 0.063 & 0.081 & 0.088 & 0.117 & 0.063 & 0.081 & 0.088 & 0.117 & 1.00 \\
Transformer & 0.021 & 0.036 & 0.037 & 0.057 & 0.021 & 0.037 & 0.075 & 0.079 & 1.38 \\
FNO & 0.125 & 0.281 & 0.270 & 0.488 & 0.127 & 0.288 & 0.474 & 0.878 & 1.80 \\
\midrule
\multicolumn{10}{@{}l}{\emph{$\Re = 3000$ ($\Rel = 25$--$50$)}} \\
UNet & 0.026 & 0.056 & 0.056 & 0.100 & 0.026 & 0.056 & 0.058 & 0.101 & 1.02 \\
Transformer & 0.050 & 0.103 & 0.108 & 0.186 & 0.051 & 0.104 & 0.110 & 0.187 & 1.01 \\
FNO & 0.383 & 0.873 & 0.724 & 1.086 & 0.385 & 0.869 & 1.032 & 1.268 & 1.17 \\
\midrule
\multicolumn{10}{@{}l}{\emph{$\Re = 6000$ ($\Rel = 44$--$101$)}} \\
UNet & 0.060 & 0.138 & 0.149 & 0.269 & 0.060 & 0.140 & 0.151 & 0.272 & 1.01 \\
Transformer & 0.058 & 0.137 & 0.160 & 0.276 & 0.058 & 0.140 & 0.162 & 0.280 & 1.01 \\
FNO & 0.539 & 1.066 & 0.904 & 1.130 & 0.539 & 1.057 & 0.996 & 1.167 & 1.03 \\
\bottomrule
\end{tabular}

\end{table}

\clearpage  %
\section{Reproducibility}
\label{app:repro}

\paragraph{Data.}
The three $\grid{128}$ data sets can be regenerated from the data generator and the seeds. Simulation $i$ at Reynolds number
$\Re$ seeds PyTorch's generator with $1000\,\Re + i$, which fixes its initial condition; App.~\ref{app:data} gives the solver
settings and the split. A simulation takes about \resMReThreeKSolverCpuHoursPerSim{} CPU-hours (medians of
\resMReOneKSolverCpuHoursPerSim{}, \resMReThreeKSolverCpuHoursPerSim{} and \resMReSixKSolverCpuHoursPerSim{} at $\Re = 1000$,
$3000$ and $6000$, on \resMReThreeKSolverRanks{} MPI ranks), so a set of 500 takes about 25,000 CPU-hours. A set holds 654\,GB in
single precision. Its 50 test simulations (65\,GB) are available on request. The two benchmark data sets of
\cref{sec:benchmarks} are public \citep{ohana2024well,li2023factformer}.

\paragraph{Code.}
Upon publication we will release the training and evaluation code, the configuration of every model in \cref{sec:mhd128,sec:benchmarks},
the evaluation scripts, the data generator, and the scripts that build the tables and figures of this paper from the result files.

\end{document}